\documentclass[preprint]{elsarticle}

\usepackage{amssymb}
\usepackage{amsmath}

\usepackage{multirow}
\usepackage{booktabs}   % For \toprule, \midrule, \bottomrule
\usepackage{caption}    % Better control over table captions
\usepackage{multirow}   % Optional: for multi-row table cells
\usepackage{siunitx}    % Optional: for numeric alignment

\usepackage{microtype} \usepackage{silence} 
\usepackage{makecell}
\usepackage{pgfplots}
\pgfplotsset{compat=newest}
\usepackage{xcolor,colortbl}
\usetikzlibrary{shapes.geometric, arrows.meta, positioning}
\usepackage{float}  % di preamble

\usepackage[english]{babel}  % Add after microtype
\usepackage{tikz}
\usetikzlibrary{arrows, arrows.meta}
\usetikzlibrary{positioning}
\usepackage{colortbl} % For \cellcolor
\usepackage{pdflscape}
\usepackage{enumitem}
\usepackage[hyphens]{url}
\usepackage{hyperref}
\hypersetup{breaklinks=true}

\usepackage{threeparttable}

\journal{Reliability Engineering \& System Safety}

\begin{document}

\begin{frontmatter}


\title{
    Beyond Foundation Models: Dimension-Aware Neural Architecture Search with Small-Data Representation Models for Cryocooler Lifetime Prediction\texorpdfstring{\tnoteref{tnote1}}{}
} %% Article title

\tnotetext[tnote1]{Accepted for publication in Reliability Engineering \& System Safety. \copyright~2026. Licensed under CC BY-NC-ND 4.0.}

% ===========================
% Authors & Affiliations (RESS / elsarticle style)
% ===========================

\author[ct-group]{Gregor Molan\corref{cor1}}
\ead{gregor@comtrade.com}
\cortext[cor1]{Corresponding author}\author[ct-group,unibo-dei]{Grafika Jati}
\author[unibo-dei]{Francesco Barchi}
\author[unibo-dei]{Andrea Acquaviva}
\author[le-tehnika]{Aljaž Osterman}
\author[ct-ai,ct-group,unibo-dei]{Martin Molan}

% --- Comtrade Group (Ljubljana) ---
\affiliation[ct-group]{
  organization={Comtrade 360 d.o.o.},
  addressline={Letališka cesta 29b},   % adjust if needed
  city={Ljubljana},
  postcode={1000},
  country={Slovenia}
}

% --- Comtrade AI (Zug) ---
\affiliation[ct-ai]{
  organization={Comtrade AI GmbH},
  addressline={Grafenauweg 8},   % adjust if needed
  city={Zug},
  postcode={6300},
  country={Switzerland}
}

% --- LE-Tehnika d.o.o. (Kranj) ---
\affiliation[le-tehnika]{
  organization={LE-Tehnika d.o.o.},
  addressline={Šuceva 27},        % keep local spelling; escape diacritics as needed
  city={Kranj},
  postcode={4000},
  country={Slovenia}
}

% --- University of Bologna, DEI (Bologna) ---
\affiliation[unibo-dei]{
  organization={Alma Mater Studiorum -- Universit\`a di Bologna, 
                Department of Electrical, Electronic, and Information Engineering (DEI)},
  addressline={Viale del Risorgimento 2}, % DEI Bologna campus; change if Cesena campus is preferred
  city={Bologna},
  postcode={40136},
  country={Italy}
}

%% Abstract
\begin{abstract}
Large-scale pretrained time-series models achieve strong results through large-scale pretraining and task-agnostic representation learning, but they rely on abundant, diverse data that industrial and scientific domains often lack. We therefore propose the FSD-RM (Family of Small-Data Representation Models) paradigm as a practical alternative for limited, domain-specific telemetry. Rather than relying on large-scale pretraining, we focus on capacity-controlled representation learning using established encoder architectures (CNN1D, LSTM, GRU, Transformer), selected for their suitability in small-data settings and interpretability.

These encoders are trained unsupervised on multivariate telemetry data and integrated into a two-stage pipeline for downstream lifetime prediction. To systematically examine architectural trade-offs under data constraints, we employ \textbf{dimension-aware neural architecture search (NAS)} to jointly optimize model capacity and input dimensionality.

Experiments on cryocooler telemetry show that the proposed approach achieves competitive predictive performance while reducing training cost and model complexity. The contribution lies in combining established representation learning techniques within a coherent, NAS-driven framework tailored to small-data regimes, with explicitly defined parameter settings and design choices. The results indicate that effective representation learning can be achieved without large-scale pretraining when appropriate inductive bias and capacity control are applied.
\end{abstract}

\noindent {\Large Graphical Abstract} \vskip12pt % as it is in elsarticle.cls
\newlength{\DSW}
\setlength{\DSW}{11.9cm}  % <--- slightly less than 12.30cm
\begin{figure}[H]
    \begin{tikzpicture}[node distance=1.2cm]
        % Nodes
        % Cryocooler data
        \node[draw, xshift=0.0cm, rectangle, rounded corners, fill=blue!10, minimum width=8.5cm, minimum height=1cm] (textRawData) {Unlabeled series of cryocoolers' telemetry data};

        % Preprocessing
        \node[align=left, draw, xshift=0.0cm, rectangle, rounded corners, below of=textRawData, yshift=-1.5cm, fill=white!5, inner sep=10pt, text width=12.30
        cm] (preprocessing) {
            \vspace{-31pt}
            {\begin{center}
                \fcolorbox{gray!90}{gray!10}{\makebox[0.5\linewidth][c]{\textbf{\textcolor{black}{Preprocessing}}}}
            \end{center} } 
            \vspace{-6pt}
            \textbf{Filtering Inv. Data}: Remove flag temp. outside -50C to 100C \\ 
            \textbf{Filtering NaN Values}: Remove seq. with missing (NaN) values \\ 
            \textbf{Normalization}: Apply Robust Scaler to all features to handle \\ 
            \makebox[\linewidth][r]{outliers and non-Gaussian distributions}
        };
        % Representation model
        \node[align=left, draw, xshift=-1.45cm, rectangle, rounded corners, below of=preprocessing, yshift=-3.4cm, fill=white!5, inner sep=10pt, text width=9.4cm] (representationModel) {
            \vspace{-31pt}
            {\begin{center}
                \fcolorbox{gray!90}{gray!10}{\makebox[0.5\linewidth][c]{\textbf{\textcolor{black}{Representation models}}}}
            \end{center} } 
            \vspace{-6pt}
            \textbf{Unsupervised Rep. Learning}: Train an encoder\\
            \makebox[\linewidth][r]{ on unlabeled time series to learn latent features} \\
            \textbf{Seq2Seq Training}: Use reconstruction loss to ensure\\
            \makebox[\linewidth][r]{embeddings capture temporal/structural patterns} \\
            \textbf{NAS Optimization}: Optimize architecture and \\ 
            \makebox[\linewidth][r]{embedding size using dimension-aware NAS} \\ 
            \textbf{Latent Embedding Extraction}: Encode all data\\
            \makebox[\linewidth][r]{into fixed-size, task-agnostic vectors}
        };

        % Downstream tasks
        \node[align=left, draw, xshift=1.45cm,
              rectangle, rounded corners,
              below of=representationModel, yshift=-4.2cm,
              fill=white!5, inner sep=10pt, text width=12.30cm] (downstreamTask) {%
            \vspace{-31pt}
            {\begin{center}
                % use fixed width \DSW instead of \linewidth
                \fcolorbox{gray!90}{gray!10}{%
                  \makebox[0.5\DSW][c]{\textbf{\textcolor{black}{Downstream tasks}}}%
                }%
            \end{center}}%
            \vspace{-6pt}
            \textbf{Reuse for Downstream Tasks}: Use embeddings for various\\
            \makebox[\DSW][r]{supervised or unsupervised downstream models} \\
            \vspace{1pt}
            \arrayrulecolor{black!50}
            \newcolumntype{P}[1]{>{\raggedright\arraybackslash}p{#1}}
            \hspace{-0.33cm}
            % table total width strictly < \DSW
            \begin{tabular}{P{5.6cm}|P{5.6cm}}
                \textbf{Binary Classification}: Use supervised classifiers to predict lifetime class
                &
                \textbf{One-Class Classification}: Use anomaly detection for one-class classification \\
            \end{tabular}\\[2pt]
            \textbf{Class Imbalance Handling}: Evaluate and mitigate the effects\\
            \makebox[\DSW][r]{of class imbalance using rebalancing and robust metrics}%
        };
        
        % Cryocooler data
        \node[draw, xshift=-0.0cm, yshift=-0.8cm, right=0.4cm of representationModel, rectangle, rounded corners, inner sep=6pt, fill=blue!10, text width=2.0cm, minimum height=2cm] (UnlabeledData) {Small \mbox{series} of \mbox{labeled} or \mbox{unlabeled} data of \textbf{new} \mbox{cryocoolers}};

        % Lifetime prediction
        \node[draw, xshift=-0.0cm, yshift=-2.0cm, rectangle, rounded corners, fill=green!10, below of=downstreamTask, minimum width=8.5cm, minimum height=1cm] (LifetimePrediction) {\textbf{Lifetime prediction of new cryocoolers}};

        % Arrows
        \draw[line width=1pt, double distance=2pt, arrows = {-Latex[fill=white,length=0pt 3 0]}, shorten >= 3mm]
            (textRawData) -- (preprocessing);
        \draw[line width=1pt, double distance=2pt, arrows = {-Latex[fill=white,length=0pt 3 0]}, shorten >= 3mm]
            (preprocessing) -- (representationModel);
        \draw[line width=1pt, double distance=2pt, arrows = {-Latex[fill=white,length=0pt 3 0]}, shorten >= 3mm]
            (representationModel) -- (downstreamTask);
        \draw[line width=1pt, double distance=2pt, arrows = {-Latex[fill=white,length=0pt 3 0]}, shorten >= 4mm]
            (UnlabeledData) -- (downstreamTask);
        \draw[line width=1pt, double distance=2pt, arrows = {-Latex[fill=white,length=0pt 3 0]}]
            (downstreamTask) -- (LifetimePrediction);
    \end{tikzpicture}
\end{figure}


%% Keywords

\end{frontmatter}

\section{Introduction}

Recent advances in time-series analysis have increasingly emphasized representation learning through large-scale, task-agnostic models. While such approaches have demonstrated strong performance, they typically rely on extensive and diverse training data, which is often unavailable in industrial and scientific applications. In domains such as satellite telemetry and cryocooler monitoring, datasets are limited, domain-specific, and costly to obtain, making direct adoption of large-scale pretraining strategies impractical. This setting motivates approaches that retain the benefits of representation learning while operating effectively under strict data constraints. In this work, we consider how representation learning can be adapted to small-data regimes by controlling model capacity, making explicit architectural choices, and systematically exploring design trade-offs.

However, existing approaches to time-series representation learning typically emphasize either increasingly complex model architectures or large-scale pretraining, with limited attention to how these choices interact with data availability. In small-data settings, overly expressive models may introduce unnecessary variance, whereas pretrained models may embed inductive biases that are poorly aligned with domain-specific signals. At the same time, commonly used architectures such as CNNs, recurrent networks, and Transformers remain effective but are often applied without systematic comparison or tuning under strict data constraints. This creates a gap in understanding how model capacity, input dimensionality, and architectural choice jointly affect representation quality in small-data regimes. To address this, we adopt a structured approach that combines a family of established encoder models with dimension-aware neural architecture search, enabling controlled exploration of design trade-offs and their impact on downstream predictive performance.

In this context, the contribution of this work is threefold. First, we formulate a small-data representation model paradigm that emphasizes controlled model capacity and task-agnostic feature learning under limited data availability. Second, we develop a family of encoder architectures based on established models (CNN1D, LSTM, GRU, Transformer) and integrate them into a two-stage pipeline for representation learning and downstream prediction. Third, we introduce a dimension-aware neural architecture search strategy that enables systematic exploration of architectural and input design choices, allowing their impact on predictive performance to be assessed in a controlled manner. The resulting framework provides a transparent and reproducible basis for studying representation learning in small-data time-series settings, with application to cryocooler lifetime prediction.

The proposed approach is motivated by cryocooler telemetry in satellite-based thermal imaging systems, where reliable lifetime prediction is essential for mission planning and system operation. The cooler is a critical component of such systems, supporting the thermal camera’s functionality for applications including environmental monitoring, target detection, and Earth observation. In this context, failures or performance degradation can have a significant operational impact, while the available telemetry data is typically limited, heterogeneous, and domain-specific.

The cooler is a critical component of thermal imaging systems, especially in satellite-based applications, where it supports the thermal camera's operation for perception tasks such as environmental monitoring, target detection, and Earth observation \cite{NAGARSHETH2018673}. Given its role in space systems, the cooler must meet stringent quality and reliability standards, as failure can compromise mission performance. Variations in cooler quality across the production pipeline require robust evaluation methods. 

Variants of Bayesian neural networks have already been employed to forecast weather-related failure risks in other safety-critical railway infrastructure, illustrating how uncertainty-aware deep models can aid in operational risk management~\cite{2022_Wang}.

\subsection{Motivation and Industrial Context}
Traditional quality assurance relies heavily on destructive lifetime testing, where coolers are operated continuously until failure to estimate durability, typically targeting a minimum operational threshold of 8,000 hours for space-grade equipment \cite{nasa2001cryocooler} \cite{olson2016microcryocooler}. This process is costly, time-consuming, and results in the destruction of high-value components.

\subsection{Challenges in Cryocooler Lifetime Prediction}
Cryocooler lifetime prediction can be formulated as a \textbf{binary classification problem}, where each unit is assigned to one of two classes based on a predefined operational lifetime threshold. Units with a lifetime below or equal to the threshold are considered to belong to the standard lifetime class, while those exceeding the threshold are classified as long-lifetime units. Classification based on a lifetime threshold is preferred over direct regression for several practical and technical reasons~\cite{KAUSIK2025100393}. First, it closely mirrors operational decision-making, where actions such as maintenance or replacement hinge on whether a component will fail before or after a specific time horizon. Second, classification models are generally more robust to noise and outliers, as regression models can be sensitive to extreme values and often exhibit high variance when precise lifetime prediction is required. Third, in many industrial contexts, labels are limited to categorical or censored data (e.g., failure before/after a given time), making regression infeasible or ill-posed. Fourth, threshold-based classification outputs are easier to interpret and integrate into automated decision-support systems. Finally, regression on single-signal time series often struggles to model the complex temporal and multivariate patterns underlying system degradation, whereas time-series classification approaches, especially those using multiple sensor inputs, are better suited to capturing these dynamics.

Despite its practical relevance, cryocooler lifetime prediction presents several fundamental challenges. These challenges include:

\begin{enumerate}
    \item \textbf{Limited labeled dataset (small-$N$ regime).}  
    Labeled data from lifetime tests is scarce due to the destructive nature of the process, which restricts the applicability of conventional supervised learning approaches.

    \item \textbf{Severe class imbalance.}  
    The dataset exhibits strong skew, with long-lifetime units significantly rarer than standard units, complicating reliable model training and evaluation.

    \item \textbf{Heterogeneous sensor modalities.}  
    Cryocooler telemetry comprises both time-domain operational signals and frequency-domain measurements with distinct statistical properties, requiring dataset-specific preprocessing and modeling strategies.

    \item \textbf{Variable-length sequences.}  
    The telemetry data consists of sequences of differing durations, introducing additional complexity for model design and representation learning.

    \item \textbf{Domain shift across test regimes.}  
    Data collected under different testing conditions exhibits distributional differences, combined with measurement noise, which can degrade generalization performance.
\end{enumerate}

These characteristics motivate the need for representation learning approaches that can exploit unlabeled data and remain robust under small-data and imbalanced conditions.

These combined challenges make the direct application of standard supervised and large-scale deep learning methods difficult in practice.

Current data acquisition and quality control practices are largely manual and based on fixed protocols, which are not only inefficient but may also fail to detect subtle anomalies~\cite{FARAHANI2025102839}. Furthermore, while multivariate time-series telemetry data from standard functional testing are abundant, they are mostly unlabeled, whereas labeled data from lifetime tests are limited. This imbalance further complicates the development of robust predictive models. Recent work on Remaining Useful Life (RUL) estimation for turbofan engines demonstrates how deep learning and attention mechanisms can improve data-driven lifetime prediction in aerospace applications~\cite{2022_Liu}.

Modern large-scale pretrained time-series models (e.g., TimesFM, UniTS, Moirai, TS2\-Vec, and transformer-based forecasting models) typically rely on large-scale pretraining over millions of heterogeneous sequences and assume access to abundant, diverse data streams. In contrast, cryocooler telemetry is characterized by low production volumes, hardware-specific signals, and datasets containing only tens to hundreds of sequences per unit, which cannot support such large-scale pretraining. Our dataset contains only 1,305 unlabeled and 95 labeled sequences. Furthermore, existing foundation models are primarily designed for forecasting tasks and do not directly address anomaly detection or lifetime classification under extreme class imbalance typical of industrial reliability datasets.

In contrast to these large-scale approaches, the framework proposed in this study operates under fundamentally different constraints. Specifically, the available dataset comprises a limited number of samples (small-$N$ regime), with only 1,305 unlabeled and 95 labeled sequences, which precludes the use of conventional large-scale pretraining strategies. Rather than aiming to replicate the scale of existing foundation models, our objective is to approximate their functional properties—such as task-agnostic representations and cross-task generalization—within a resource-constrained, domain-specific setting.

To address these limitations, we adopt a \textbf{modular machine learning pipeline} in which data flows through a sequence of preprocessing, representation learning, and downstream modeling stages~\cite{2023-ModiTowards}. The core of this approach is a \textbf{capacity-scalable representation framework}, referred to as a small-data representation model paradigm, in which a family of encoder models is trained in an unsupervised sequence-to-sequence setting to learn general-purpose representations from unlabeled telemetry data~\cite{2024-Schneider}. These representations are subsequently reused by lightweight downstream models~\cite{2024-pmlr-v235-qiu24d} for lifetime classification and anomaly detection, enabling robust predictions even under severe data scarcity and class imbalance.

\subsection{Contributions of This Work}
\begin{figure}[H]
    \centering
    \includegraphics[width=0.8\textwidth]{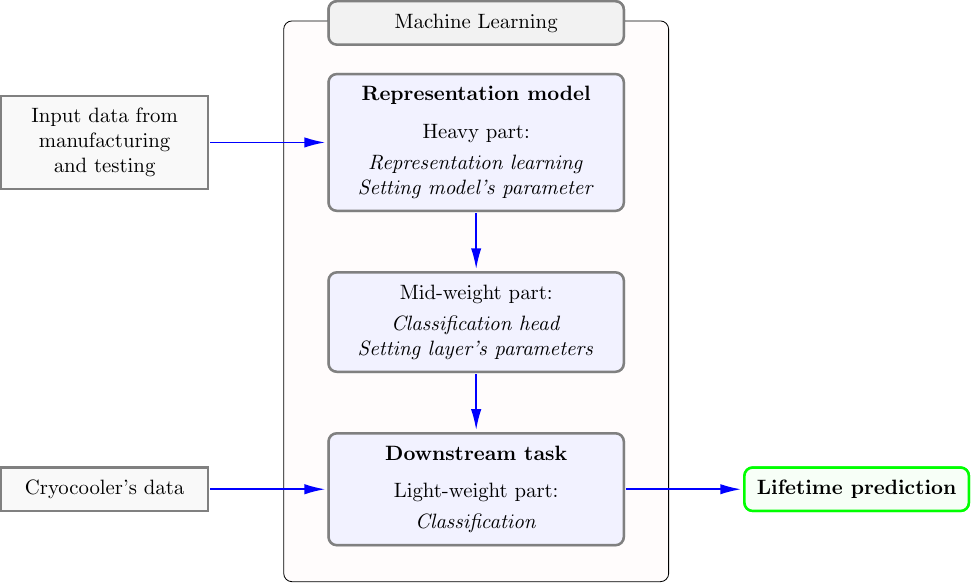}
    \caption{An architecture of a small-data representation model and downstream tasks.}
    \label{fig:model_parts}
\end{figure}

We present a novel approach to time series lifetime prediction in low-label industrial settings, with the following core contributions:
\begin{enumerate}
    \item \textbf{Small-Data Representation Modeling for Cryocooler Satellite Te\-le\-metry.}
    We introduce an FSD-RM paradigm specifically designed for cryocooler satellite telemetry, demonstrating that generalization can emerge under extremely small-data constraints, far below the scale required by existing large-scale pretrained time-series models.
    
    \item \textbf{Capacity-Scalable Family of Small-Data Representation Models.}
    We introduce the FSD-RM, a Family of Small-Data Representation Model encoders—CNN1D, LSTM, GRU, and Transformer—parameterized by embedding capacity. This FSD-RM enables controlled scaling of representational complexity, allowing downstream adaptation to varying data regimes, sensor behaviors, and operational constraints in aerospace telemetry.

    \item \textbf{Cross-Task Transferability of Learned Representations.} 
    We demonstrate that a single pretrained representation supports multiple heterogeneous downstream tasks—binary fault classification and one-class anomaly detection—without retraining the encoder. This cross-task generality confirms the task-agnostic nature expected from a proper model in satellite telemetry monitoring.

    \item \textbf{Multi-Regime Robustness Under Data Scarcity and Class Rarity.}
    We assess the proposed representations under progressively reduced downstream training regimes, spanning from full-data availability down to ultra-low-sample scenarios that reflect real cryocooler manufacturing constraints. These subsets exhibit both a limited sample availability and increasingly skewed class distributions, mimicking degradation-stage rarity in operational telemetry.

    \item \textbf{Dimension-Aware NAS for Optimal Embedding Capacity.}
    We introduce a lightweight NAS strategy (\textbf{da-NAS}) that selects the optimal embedding dimension using a progressive dimension schedule, a Beat-Lower-Dimension improvement rule, and a simple dimension-aware early stopping criterion. This capacity-oriented search differs from prior NAS methods focused on depth/width scaling.
    
    \item \textbf{System-Level Design Beyond Single-Model Optimization.}  
    Unlike conventional approaches that focus on improving individual model architectures, our contribution lies in a system-level framework that integrates family-based representation learning, capacity scaling, cross-task reuse, and dimension-aware NAS into a unified pipeline for small-data industrial environments.
\end{enumerate}
As shown in Figure~\ref{fig:model_parts}, we propose a two-stage learning pipeline, from the representation model to downstream tasks, that extracts meaningful representations from cryocooler telemetry and enables reliable downstream lifetime prediction.

To the best of our knowledge, this is among the first systematic works that propose a family of pretrained time-series encoders spanning embedding dimensions from 2 to 512 and designed to support two fundamentally distinct downstream tasks: supervised binary classification—separating cryocoolers operating below the lifetime threshold (class 0) from those exceeding it (class 1)—and unsupervised one-class classification, where the encoder is trained solely on class-0 telemetry to detect anomalous high-lifetime behavior during testing. This encoder family offers flexibility, allowing model capacity to be matched to data availability and task complexity. We further validate the approach on operational cryocooler telemetry and show that the learned representations capture degradation characteristics, enabling lifetime-related quality assessment from short test segments without performing complete lifetime testing and supporting early-failure detection in satellite cryocooler manufacturing.

\section{Problem Characteristics}

A key challenge in industrial and aerospace telemetry is that datasets are inherently small, highly imbalanced, and domain-specific. Unlike large-scale internet corpora that enable GPT-style large-scale pretrained models, cryocooler and space-system manufacturing typically operate under low production volumes, resulting in small but highly specialized datasets. This makes it impractical to rely on large pretrained time-series models or general-purpose pretrained models, which require millions of samples and offer limited transferability to niche sensor modalities.

What is missing in the literature is an approach that provides cross-task generalization, a reusable representation, and classifier-agnostic embeddings—without assuming access to large datasets. In other words, a framework that works in real industrial conditions:
few samples, strong imbalance, and highly domain-specific signals.

To address this gap, we introduce a Family of Small-Data Representation Models (\textbf{FSD-RM}): a collection of capacity-scaled encoders (CNN1D, LSTM, GRU, Transformer) with systematically varied embedding dimensions. Instead of relying on a single universal model, this family enables dynamic selection of model capacity depending on the downstream requirement—whether the task is binary classification, one-class anomaly detection, or small-sample learning. This capacity scaling and cross-task consistency provide the functional characteristics of FSD-RM while remaining feasible for domains where conventional large-scale pretraining is impossible.

Our “family-based” formulation primarily treats representation models as a single, large, monolithic representation, whereas our work demonstrates that a set of smaller, specialization-aware representation model variants can collectively provide FSD-RM properties under severe data constraints. The proposed approach is therefore directly aligned with the operational needs of cryocooler manufacturing and similar space-industry pipelines: enabling downstream flexibility while respecting fundamental dataset limitations.

\section{Related Work}
\subsection{Aerospace Reliability and Remaining Useful Life Estimation}
Rosero proposed another study related to the Remaining Useful Life~\cite{aerospace9060309}. The study presents a hybrid PHM approach that estimates the RUL of aircraft cooling units by combining physics-based health indicators with machine learning, leveraging time-frequency features such as the Hilbert spectrum to detect degradation.

Recent work on turbofan engines demonstrates the effectiveness of attention-based deep learning architectures for RUL estimation in aerospace systems, using double-attention mechanisms in the C-MAPSS benchmark dataset~\cite{2022_Liu}.
In addition, Chen et al. show that jointly modeling aleatoric, epistemic, data-missing, and semantic uncertainties via a Bayesian BiGRU architecture can significantly improve the reliability of RUL prediction~\cite{2025_Chen}.

By leveraging run-to-failure data instead of raw sensor signals, the method improves accuracy in identifying abnormal degradation and reduces prediction error. Existing studies use sensors during equipment runtime—rather than relying on quality assurance checks or machine part inspections that do not involve operating the equipment under real conditions. Meanwhile, quality assurance for machine parts immediately after production has not yet been addressed.

\subsection{General RUL and Predictive Maintenance}
In modern discrete manufacturing industries, ensuring product quality, detecting faults early, and maintaining equipment reliability are critical tasks supported by effective quality control, inspection, and predictive maintenance strategies. Recognizing the increasing role of Artificial Intelligence (AI) in these areas, Kausik et al. conducted a comprehensive review that categorizes AI applications into key functional domains, including predictive maintenance, quality inspection, process optimization, and autonomous operations~\cite{KAUSIK2025100393}. The review analyzes real-world case studies, emphasizing that most datasets are sourced from sensor data collected during equipment runtime, rather than from synthetic data or post-production inspections. The surveyed technical approaches encompass a variety of machine learning and deep learning models, including CNNs, LSTMs, autoencoders, Random Forest, SVMs, and hybrid methods that employ time-frequency representations.

For time-series data in smart manufacturing, Farahani et al. summarize various use cases, including fault detection, tool wear monitoring, process phase classification, and post-production quality prediction, in which sensor signals are continuously recorded during equipment operation~\cite{FARAHANI2025102839}. These time-series classification tasks are essential for enabling real-time decisions, and the authors benchmark state-of-the-art machine learning models—including InceptionTime, ROCKET, and HIVE-COTE—across multiple datasets to evaluate their effectiveness in these industrial scenarios.

Tašći proposed a machine learning-based system to predict the Remaining Useful Life (RUL), using a hybrid approach that combines data filtering, autoencoder-based feature engineering, clustering, and regression models (RF, XGBoost, MLP, SVR)~\cite{TASCI2023109566}. The system is designed for predictive maintenance in real-world production lines, using historical operational data collected from IoT sensors during equipment runtime.

Classical supervised learning approaches have been widely applied to reliability and predictive maintenance, providing established baselines for understanding lifetime prediction. Ouadah et al. study the systematic selection of supervised machine learning algorithms for predictive maintenance, comparing Random Forest, Decision Tree, and k-Nearest Neighbors for their effectiveness in lifetime and reliability prediction tasks~\cite{Ouadah2022}.

Felsberger et al. formulate reliability assessment as a Bayesian regression problem to derive predictive models of reliability metrics, with applications to operational particle accelerator equipment at CERN~\cite{Felsberger2018}.

Su and Chiang combine finite element analysis (FEA) with supervised machine learning methods, such as Kernel Ridge Regression, to predict the reliability life of wafer-level packaging in electronic components~\cite{Su2022}.

% Gregor: Add RESS citation
Recent studies on wind-induced floater intrusion risk in railway overhead contact lines indicate that Bayesian neural networks can provide calibrated, uncertainty-aware risk predictions even with small and imbalanced datasets~\cite{2022_Wang}.

While these classical methods establish important baselines, they rely on explicit feature engineering and are limited in scenarios with extreme class imbalance and scarce labeled data, motivating the need for representation learning approaches that leverage unlabeled data.

\subsection{Time-Series Representation Learning and Foundation-Style Models}
\subsubsection{Foundation Models and Representation Learning for Reliability}
In recent years, foundation model approaches have expanded beyond NLP to include time-series domains, enabling unsupervised learning of representations from unlabeled data. Schneider et al. define foundation models as large pretrained encoders that develop general-purpose, reusable representations~\cite{2024-Schneider}.

In reliability and warranty contexts, Kim et al. propose a weighted temporal convolutional autoencoder (WTCAE) designed to predict field reliability and claim counts from limited initial warranty data~\cite{2025_Kim-Limited_claim_data}. The WTCAE performs better at predicting warranty from short-term claim data, where traditional lifetime distribution methods often fail. This research is especially relevant because it addresses a similar challenge: predicting long-term reliability with small sample sizes and imbalanced data. Likewise, structured latent-space autoencoders have been used to learn compact representations of sensor data for health monitoring. These representation-learning methods inspire our FSD-RM design, which extends encoder-decoder approaches to the field of aerospace cryocoolers.

\subsubsection{Recurrent and Adaptive Architectures}
Recent work on industrial health monitoring has proposed specialized recurrent architectures for multivariate time series. Huang et al. (2025) introduce a multivariate time-series adaptive GRU (MTS-AdaGRU) with transfer learning designed for coal mill systems. The MTS-AdaGRU combines temporal distribution characterization, factorized temporal mixing, and temporal distribution matching to address temporal covariate shift and redundancy in multivariate sensor data~\cite{2025_Congzhi}. They define a health degree metric based on Jensen–Rényi divergence between actual values and model outputs, providing a scalar indicator of deviation from normal behavior. In real plant data, MTS-AdaGRU achieves significantly smaller and more focused errors compared to baselines such as GRU, TCN, Transformer, and LSTNet across multiple verification sets, demonstrating strong generalization and robustness under changing operating conditions.

Our approach addresses a related but distinct challenge. While MTS-AdaGRU optimizes a single recurrent architecture with domain-shift adaptation, our work systematically evaluates a family of heterogeneous encoder architectures (CNN1D, LSTM, GRU, Transformer) with varying capacities, coupled with a dimension-aware neural architecture search (da-NAS) strategy. Instead of adapting a single model to distributional shifts, we enable capacity matching to data availability—a critical requirement for aerospace cryocooler manufacturing, where extremely limited labeled samples and severe class imbalance are inherent constraints. This capacity-scalable family design, combined with cross-task validation on both binary classification and one-class anomaly detection, constitutes a unique contribution to small-data reliability prediction in specialized, low-volume domains.

\section{Problem Specification}

\subsection{Positive-Unlabeled Imbalanced Multivariate Time Series}

This study uses an in-house dataset of multivariate time-series signals acquired from cryocooler production tests. A total of \textbf{1,305 unlabeled sequences} are available for representation learning, while \textbf{95 sequences} include ground-truth lifetime annotations and are used for downstream evaluation.

The downstream task formulates lifetime estimation as a binary classification problem defined by a threshold \(T\). Units with lifetime \(\leq T\) are labeled as \textbf{class 0 (standard lifetime)}, whereas those exceeding \(T\) are labeled as \textbf{class 1 (long lifetime units)}. Three thresholds are evaluated to reflect typical operational regimes in a cryogenic cooler qualification:

\begin{description}[itemsep=0pt, parsep=0pt, topsep=2pt]
    \item[10,000-hour threshold:] 63 samples in class 0 and 32 samples in class 1.
    \item[15,000-hour threshold:] 73 samples in class 0 and 22 samples in class 1.
    \item[20,000-hour threshold:] 84 samples in class 0 and 11 samples in class 1.
\end{description}

\begin{figure}[H]
    \centering
    \includegraphics[width=1\textwidth]{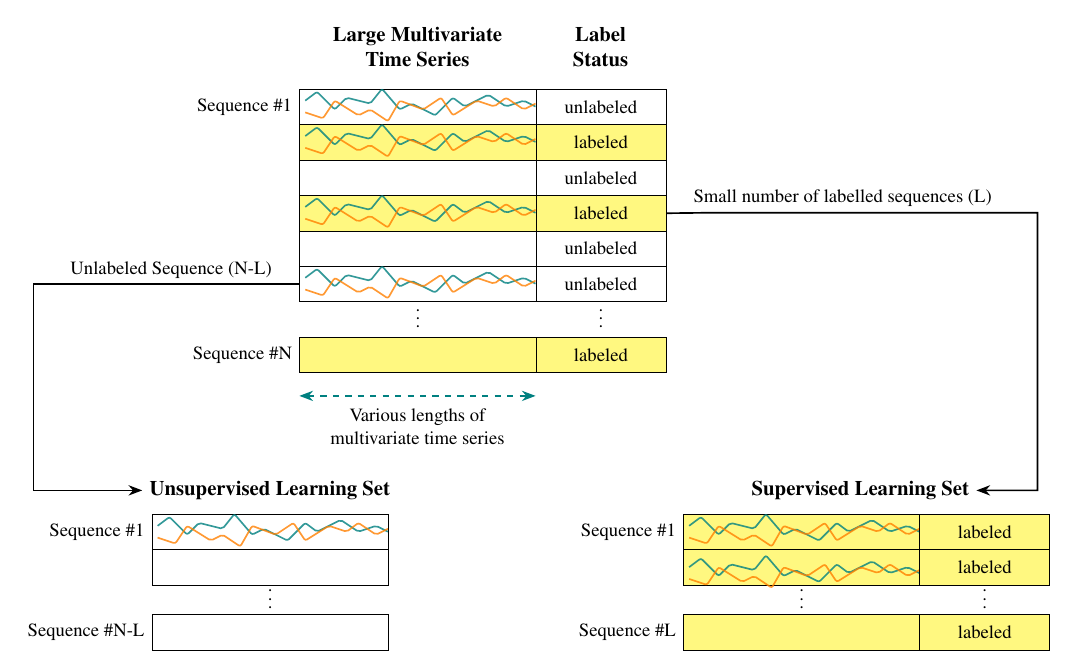}
    \caption{Challenge on Positive-Unlabeled Imbalanced Multivariate Time Series.}
    \label{fig:data_problem}
\end{figure}

These distributions exhibit progressively increasing imbalance from \textbf{mild (10k)} to \textbf{medium (15k)} and \textbf{severe (20k)} class skew, reflecting realistic manufacturing statistics in which long lifetime units are substantially rarer than standard ones. This setting motivates the need for a representation approach capable of handling \textbf{limited labeled data}, \textbf{positive-unlabeled conditions}, and \textbf{extreme class rarity}, all intrinsic to satellite cryocooler telemetry. Figure~\ref{fig:data_problem} visually presents the challenges of Positive-Unlabeled and imbalanced multivariate time series.

\subsection{Definition of Cryocooler Time Series Representation Model}

We define a Family of Small-Data Representation Models (FSD-RM) as a set of heterogeneous encoders—CNN1D, LSTM, GRU, and Transformer—pretrained to learn a generic representation of multivariate time-series data. These models share the same objective but differ in architectural principles, capacity, and inductive bias. Furthermore, we approximate the properties of large-scale representation models for our FSD-RM, enabling reusable representations, cross-task generalization, and independence from downstream tasks.

Among the FSD-RM candidates, we identify a representation model as the encoder that demonstrates the most consistent representation quality across all downstream tasks, including binary classification and one-class anomaly detection. The representation model is therefore considered the universal backbone for representations in our framework.

\section{Proposed Method: Family of Small-Data Representation Models for Cryocooler Telemetry} 
\noindent\textbf{Notation.} Let $B$ denote the batch size, $T$ the sequence length, and $F$ the number of input features. The input time series is denoted as $X \in \mathbb{R}^{B \times T \times F}$, and $X' \in \mathbb{R}^{B \times F \times T}$ represents its channel-first permutation. The number of convolutional layers is denoted by $n_{\text{conv}}$, and $d_{\text{emb}}$ denotes the embedding dimension (ranging from 2 to 512). Intermediate hidden representations are denoted by $H^{(l)}$, while $Z$ represents the latent embedding and $\hat{X}$ the reconstructed output.

Additional model-specific parameters include trainable projection matrices and bias terms, denoted by $\mathbf{W}_{\text{emb}}$, $\mathbf{b}_{\text{emb}}$, $\mathbf{W}_{\text{in}}$, and $\mathbf{b}_{\text{in}}$. In the Transformer encoder, $\mathbf{P}$ denotes positional encoding, $\mathbf{M}$ the encoder memory representation, $\boldsymbol{\alpha}$ the attention weights, and $\mathbf{w}_{\text{att}}$ the trainable attention vector. In the da-NAS formulation, $T_{D_k}$ denotes the best validation-loss target at embedding dimension $D_k$, $L_{\text{val}}$ denotes validation loss, $L_D^{\*}$ denotes the best validation loss at dimension $D$, $L_t$ denotes the validation loss at trial $t$, $L^{*}$ denotes the current best loss, and $\delta$ denotes the relative improvement threshold. BLD refers to the Beat-Lower-Dimension criterion, and CDSP denotes the Cross-Dimensional Stop Policy.

\begin{figure}[H]
    \centering
    \includegraphics[width=1\textwidth]{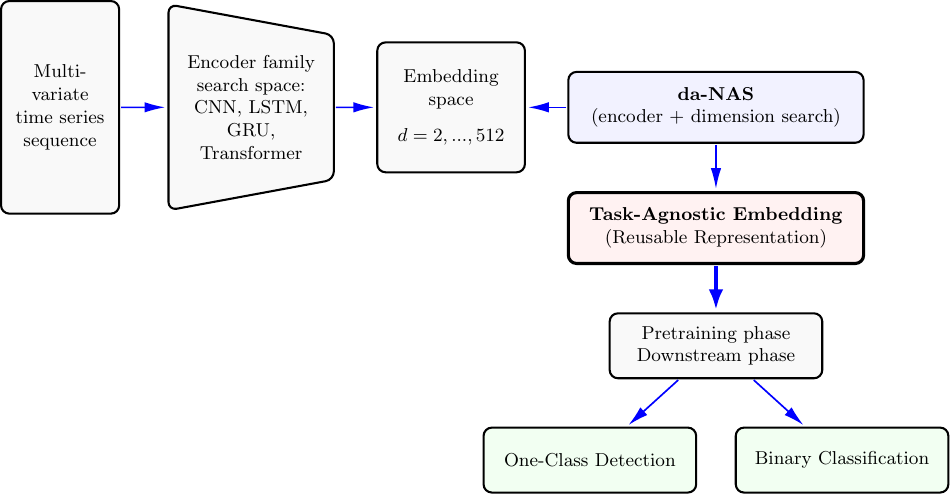}
    \caption{Proposed small-data representation model for cryocooler lifetime prediction model.}
    \label{fig:proposed_method}
\end{figure}

Figure~\ref{fig:proposed_method} illustrates the overall architecture of the proposed approach. We adopt an FSD-RM paradigm where an encoder--decoder architecture is pretrained using multivariate cryocooler telemetry time series in an unsupervised sequence-to-sequence reconstruction setting. The goal is to learn embedding representations that are compact, generalizable, and reusable across downstream tasks.

Unlike prior work that focuses on a single model, our approach introduces a \textbf{Family of Small-Data Representation Models (FSD-RM)} — a set of pretrained encoders based on four architectures (CNN1D, LSTM, GRU, Transformer) with scalable embedding capacities. This design allows us to match model complexity to data availability and downstream task requirements, which is critical in small-sample domains like satellite cryocooler manufacturing.

The downstream tasks are decoupled from the encoder training and applied to the learned embeddings:
\begin{description}
    \item[Binary Classification:] Predict whether a cryocooler unit will fail before or after a predefined operational lifetime threshold (e.g., 10,000 hours).
    \item[Anomaly Detection (One-Class Classification):] Train on only normal (st\-andard lifetime) samples and detect outliers that exhibit abnormally long behavior as potential quality deviations.
\end{description}

By training a set of encoders across a wide range of embedding dimensions (from 2D to 512D), we enable capacity-scaled representation learning that supports performance-efficiency tradeoffs and robustness under data scarcity. This method is particularly suited for the space manufacturing domain, where the acquisition of labeled failure data is costly and time-consuming.

A related research approach involves designing normal‐behavior models that explicitly handle shifts in temporal distribution in industrial time series. In coal mill health monitoring, Huang et al. combine an adaptive GRU with techniques such as temporal distribution characterization, factorized temporal mixing, and temporal distribution matching to learn features that remain consistent across different operating periods and are resistant to covariate shift. Our time-series representation models pursue a similar objective of creating regime-robust representations, but accomplish this by pretraining various encoder architectures (CNN1D, LSTM, GRU, Transformer) on reconstruction tasks and then selecting capacity through dimension-aware NAS, rather than integrating adaptation mechanisms within a single recurrent model~\cite{2025_Congzhi}.

\subsection{Data Acquisition}

\begin{figure}[H]
    \centering
    \includegraphics[width=\textwidth]{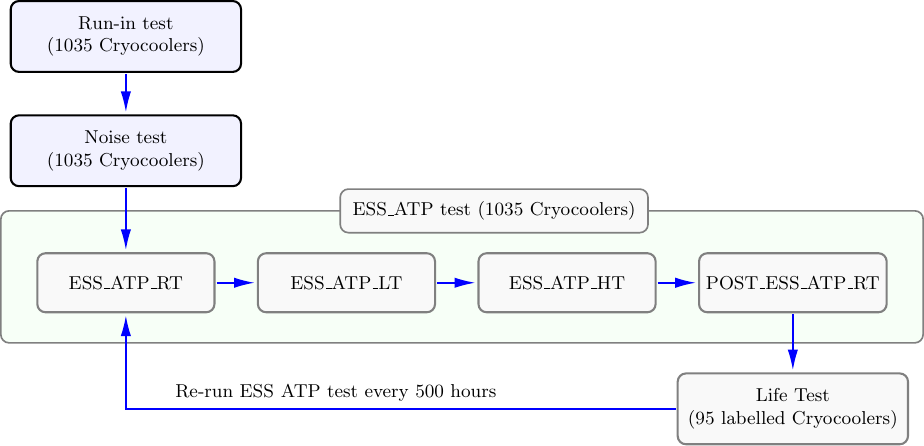}
     \caption{Data acquisition process.}
    \label{fig:data_acq}
\end{figure}
High-level design of our data acquisition process is presented in Figure~\ref{fig:data_acq}. The cryocooler's post-production testing process consists of several structured phases. Run-in is the initial phase, lasting 150 hours with data collected every minute, aimed at stabilizing device performance. Next, the Noise Test runs for 15 minutes with a 1-second resolution, measuring vibration frequencies; it is performed on all coolers. The ESS (Environmental Stress Screening) includes four phases: RT (Room Temperature), LT (Low Temperature, -40°C), HT (High Temperature, 71°C), and Post ESS RT, each separated by stabilization periods. Data is recorded at 0.5-second intervals. If successful, the device proceeds to the ESS\_ATP (Acceptance Test Procedure), which mirrors the ESS sequence and is conducted before and after the life test. Finally, the Life Test involves continuous operation, with ESS tests repeated every 500 hours to verify continued reliability before resuming operation.

\begin{figure}[H]
    \centering
    \includegraphics[width=\textwidth]{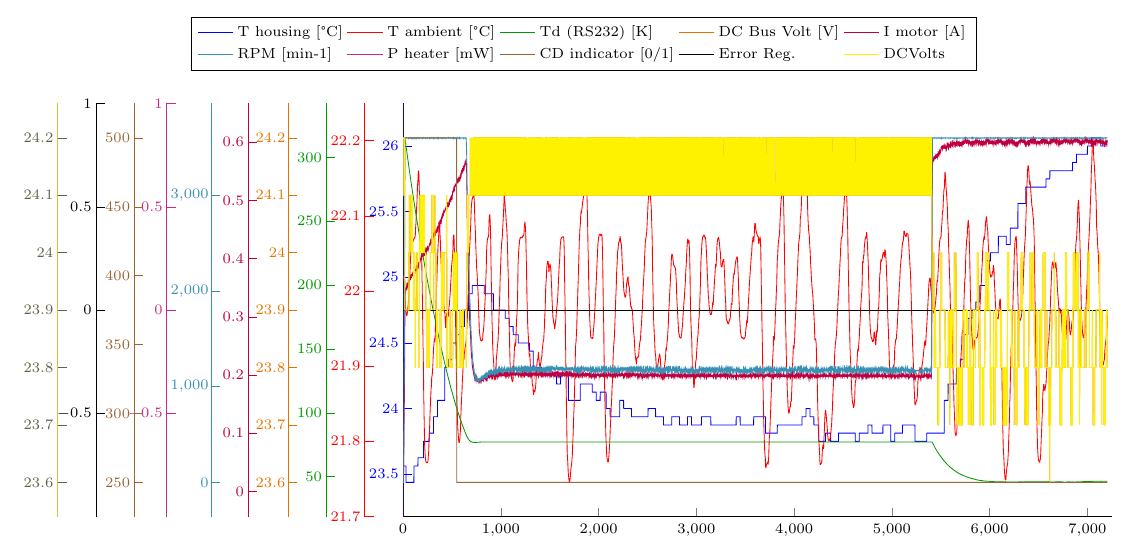}
    \caption{Sample of a Multivariate time series taken from ESS\_ATP\_RT test.}
    \label{fig:data_rt}
\end{figure}

The cooler telemetry dataset is a large-scale multivariate dataset collected from industrial cryocooler systems. It includes a wide range of features that capture both environmental and operational characteristics, such as T\_housing, T\_ambient, DC\_Bus, I\_motor, P\_heater, DCVolts, as well as high-level control variables like Error\_Reg, CD, and RPM, as shown in the legend. For the ESS\_ATP tests—including ESS\_ATP\_LT, ESS\_ATP\_HT, ESS\_ATP\_RT, a consistent set of 10 core telemetry features is used to monitor system behavior over time. A sample of the ESS\_ATP\_RT test is shown in Figure~\ref{fig:data_rt}.

The NoiseTest dataset captures high-frequency signals, specifically vibration and acoustic spectra. It comprises 33 frequency-domain features, including spectral power across multiple frequency bands (20 Hz to 20,000 Hz), RPM, and total band power. These features are not analyzed individually; instead, they are jointly used as input to the representation learning models. This enables the extraction of compact, task-agnostic embeddings that capture the overall spectral characteristics relevant for lifetime prediction and anomaly detection.

The cryocooler telemetry dataset poses several modeling challenges due to its structural and statistical properties. First, it exhibits feature heterogeneity across test types: ESS\_ATP datasets contain time-domain telemetry signals (e.g., voltage, temperature, RPM), while the NoiseTest dataset focuses on frequency-domain features (e.g., spectral band power). This difference requires dataset-specific preprocessing strategies rather than a unified feature representation. Second, the dataset contains variable-length time series, as each test is recorded over a different duration, leading to sequences of inconsistent lengths that complicate model design. Third, class imbalance and sparsity of anomalous events—such as noise or faults—can impair classifier performance if not addressed. Fourth, domain shift caused by different test conditions (e.g., operational settings, data source) limits model generalizability across subsets. Finally, sensor noise and measurement uncertainty can introduce irregularities in the data, requiring filtering or robust representation learning techniques.

Importantly, time-domain and frequency-domain telemetry are not fused into a single unified input representation. Instead, each modality is processed using a modality-specific preprocessing pipeline tailored to its statistical properties. The representation learning models are trained separately on each dataset, ensuring that encoder embeddings capture modality-consistent structures without introducing cross-modal interference. This design reflects the operational separation of test regimes and preserves the interpretability of learned representations.

\paragraph{Organization of Time- and Frequency-Domain Features}
The cryocooler telemetry consists of two distinct modalities: (i) time-domain operational signals (e.g., temperature, voltage, RPM) and (ii) frequency-domain spectral features obtained from NoiseTest measurements. Due to their distinct statistical properties—such as sampling rate, temporal structure, and distributional characteristics—these modalities are processed independently and not concatenated at the feature level. 

For the time-domain data, each sequence is represented as a multivariate temporal tensor $X \in \mathbb{R}^{B \times T \times F}$, where $T$ corresponds to the temporal dimension and $F$ to the set of sensor channels. These sequences are directly fed into the encoder architectures (CNN1D, LSTM, GRU, Transformer), preserving temporal order and enabling sequential representation learning.

In contrast, the frequency-domain data is organized as fixed-length vectors or short sequences derived from spectral measurements. These features are treated as structured multivariate inputs, where each frequency band corresponds to a channel. Although they do not represent long temporal dynamics, they are processed using the same encoder framework by treating the frequency axis as a structured input dimension, enabling models to learn compact representations of spectral patterns.

By maintaining separate preprocessing pipelines and training independent representation models for each modality, the proposed framework avoids bias introduced by heterogeneous feature fusion and ensures that each encoder captures modality-specific structure. The resulting embeddings are therefore consistent within each modality while remaining compatible with the downstream classification and anomaly detection tasks.

\subsection{Sequence Preprocessing}
\subsubsection{Filtering Invalid Data}
An important step in the preprocessing pipeline involves validating the temperature data recorded in the column \texttt{T housing [°C]}, which represents the housing temperature in degrees Celsius. The process first checks whether this column exists in the input time series sequence. If it is present, the code evaluates whether any recorded values fall outside a physically reasonable range, specifically below $-50^\circ$C or above $100^\circ$C. Such values may indicate sensor malfunction, corrupted measurements, or outliers, and can be flagged for further handling or removal. 

\subsubsection{Filtering NaN Value}
In addition to range validation, the preprocessing also includes a step to remove sequences that contain missing values (\texttt{NaN}). This is a necessary operation because most machine learning models and analytical methods cannot process incomplete data. Dropping rows with \texttt{NaN} ensures that the dataset used for inference or training is clean, consistent, and reliable.

\subsubsection{Normalization}
Given that our multivariate time series data contains non-Gaussian distributions and is susceptible to outlier values, especially during abnormal test conditions, Robust Scaler provided more stable training and reduced sensitivity to extreme values, leading to better overall reconstruction performance. We selected Robust Scaler, as it demonstrated greater resilience to outliers and non-Gaussian feature distributions commonly observed in our sensor data, resulting in more stable training dynamics. 

\subsection{Representation Learning Framework}
\label{subsec:Representation_learning_Framework}
The proposed framework employs an unsupervised sequence-to-sequence representation learning architecture designed to encode multivariate time-series data into a compact latent embedding. 
Similar encoder–decoder paradigms have been studied in warranty and field-reliability contexts, where Kim et al. use a weighted temporal convolutional autoencoder (WTCAE) to learn from limited claim data and forecast long-term field reliability despite short-term, imbalanced observations~\cite{2025_Kim-Limited_claim_data}.
The objective is to map the complete temporal sequence into a low-dimensional latent space that preserves salient temporal dependencies, structural patterns, and feature dynamics. The input multivariate sequence is processed through an encoder–decoder pipeline, where the encoder extracts hierarchical temporal features, and the decoder reconstructs the input sequence to enforce meaningful latent representations.

To enable downstream tasks to operate on fixed-size inputs, the learned temporal features are aggregated into a global latent representation via temporal pooling. This aggregation step, as defined in Equation~(\ref{eq:Seq2SeqCNN1D_Encoder_Z}), transforms variable-length sequence representations into a compact embedding vector that captures the input's overall temporal dynamics. While this operation is explicitly defined in the Seq2SeqCNN1D architecture, the same principle of temporal aggregation is consistently applied across all encoder variants in the proposed framework.

The encoder module is designed as a pluggable architecture, allowing different backbone models depending on the desired temporal modeling capacity. In this work, we investigate four variants of the encoder: CNN1D, LSTM, GRU, and Transformer,
reflecting architectures that have also proven effective for Remaining Useful Life (RUL) and reliability prediction in aerospace systems~\cite{2022_Liu}.
The resulting latent embedding serves as a compact, task-agnostic representation that can be directly utilized for downstream tasks such as classification or anomaly detection.

% \newpage
\subsubsection{Seq2SeqCNN1D: Temporal Convolutional Autoencoder}
\label{subsec:cnn1d}

As part of the model search space, we include a lightweight convolutional encoder–decoder, \textbf{Seq2SeqCNN1D}, designed for sequence reconstruction and embedding extraction from multivariate time series.

\paragraph{Motivation}
Temporal convolutional networks provide an efficient alternative to recurrent architectures for modeling local temporal dependencies. Seq2SeqCNN1D leverages stacked 1D convolutions to produce compact latent representations suitable for reconstruction-based self-supervision and downstream tasks.

\paragraph{Architecture}
Given an input sequence $\mathbf{X} \in \mathbb{R}^{B \times T \times F}$, where $B$ is the batch size, $T$ is the sequence length, and $F$ is the number of features, the model consists of an encoder and a decoder.

\paragraph{Encoder}
The input is first permuted to $\mathbf{X}' \in \mathbb{R}^{B \times F \times T}$. A stack of $n_{\text{conv}}$ convolutional layers transforms the representation according to Equation~(\ref{eq:Seq2SeqCNN1D_Encoder}):
\begin{equation}
\label{eq:Seq2SeqCNN1D_Encoder}
    \mathbf{H}^{(l)} = \text{Dropout}\bigl(\text{ReLU}(\text{CNN1D}(\mathbf{H}^{(l-1)}))\bigr), \quad l = 1,\dots,n_{\text{conv}}.
\end{equation}
This transformation progressively extracts hierarchical temporal features from the input sequence.

The encoder output $\mathbf{Z} \in \mathbb{R}^{B \times d_{\text{emb}} \times T}$ is aggregated into a fixed-dimensional representation through global average pooling, as defined in Equation~(\ref{eq:Seq2SeqCNN1D_Encoder_Z}):
\begin{equation}
\label{eq:Seq2SeqCNN1D_Encoder_Z}
    \mathbf{z}_{\text{global}} = \frac{1}{T}\sum_{t=1}^{T}\mathbf{Z}_t \in \mathbb{R}^{B \times d_{\text{emb}}}.
\end{equation}
This operation produces a compact embedding independent of sequence length, as formally defined in Equation~(\ref{eq:Seq2SeqCNN1D_Encoder_Z}).

\paragraph{Decoder}
The reconstruction is obtained by projecting the latent representation back to the original feature space using a $1 \times 1$ convolution:
\begin{equation}
\label{eq:Seq2SeqCNN1D_Decoder}
    \widehat{\mathbf{X}} = \text{CNN1D}_{1\times1}(\mathbf{Z}), \quad \widehat{\mathbf{X}} \in \mathbb{R}^{B \times F \times T}.
\end{equation}

\paragraph{Outputs}
The model produces the global embedding $\mathbf{z}_{\text{global}} \in \mathbb{R}^{B \times d_{\text{emb}}}$ and the reconstructed sequence $\widehat{\mathbf{X}} \in \mathbb{R}^{B \times T \times F}$, as formally defined in Equation~(\ref{eq:Seq2SeqCNN1D_Decoder}).

The transformation defined in Equation~(\ref{eq:Seq2SeqCNN1D_Encoder}) extracts hierarchical temporal features.

\vspace{0.5em}

\subsubsection{Seq2SeqLSTM}
\label{subsec:lstm}

The \textbf{Seq2SeqLSTM} architecture employs recurrent layers to capture long-range dependencies and uses sequence reconstruction as a self-supervised learning objective.

\paragraph{Motivation}
LSTM networks effectively model non-stationary temporal dynamics and long-term dependencies, enabling the extraction of embeddings that summarize temporal evolution for downstream tasks.

\paragraph{Architecture}
For $\mathbf{X} \in \mathbb{R}^{B \times T \times F}$, the model consists of an encoder and a decoder.

\paragraph{Encoder}
The input sequence is processed by stacked LSTM layers:
\begin{equation}
\label{eq:LSTM_Encoder}
    (\mathbf{H}, (\mathbf{h}_T,\mathbf{c}_T)) = \text{LSTM}_{\text{enc}}(\mathbf{X}),
\end{equation}
where $\mathbf{H} \in \mathbb{R}^{B \times T \times H}$ is obtained as defined in Equation~(\ref{eq:LSTM_Encoder}).

The hidden states are projected into the embedding space and aggregated as follows:
\begin{subequations}
\begin{align}
    \label{eq:LSTM_Encoder_Z}
    \mathbf{Z} &= \mathbf{H}\mathbf{W}_{\text{emb}}^\top + \mathbf{b}_{\text{emb}}, \\
    \label{eq:LSTM_Encoder_z_global}
    \mathbf{z}_{\text{global}} &= \frac{1}{T}\sum_{t=1}^{T}\mathbf{Z}_{:,t,:}.
\end{align}
\end{subequations}
These operations produce a fixed-dimensional representation of the sequence.

\paragraph{Decoder}
The decoder reconstructs the sequence using an LSTM initialized with the encoder states:
\begin{equation}
\label{eq:LSTM_Dencoder}
    \mathbf{H}_{\text{dec}} = \text{LSTM}_{\text{dec}}(\mathbf{Z};\mathbf{h}_T,\mathbf{c}_T).
\end{equation}
A linear projection is subsequently applied to obtain $\widehat{\mathbf{X}} \in \mathbb{R}^{B \times T \times F}$, based on the decoder formulation in Equation~(\ref{eq:LSTM_Dencoder}).

\paragraph{Outputs}
The outputs consist of the reconstructed sequence $\widehat{\mathbf{X}}$ and the embedding $\mathbf{z}_{\text{global}}$.

\vspace{0.5em}

\subsubsection{Seq2SeqGRU}
\label{subsec:gru}

The \textbf{Seq2SeqGRU} architecture provides a parameter-efficient recurrent alternative while retaining the ability to model long-range dependencies.

\paragraph{Motivation}
GRUs reduce model complexity compared to LSTMs, making them well suited for small-data scenarios while maintaining effective temporal modeling.

\paragraph{Architecture}
For $\mathbf{X} \in \mathbb{R}^{B \times T \times F}$, the model consists of an encoder and a decoder.

\paragraph{Encoder}
The sequence is processed using stacked GRU layers:
\begin{equation}
\label{eq:Seq2SeqGRU_Encoder}
    (\mathbf{H},\mathbf{h}_T) = \text{GRU}_{\text{enc}}(\mathbf{X}).
\end{equation}
The hidden states $\mathbf{H}$, obtained as defined in Equation~(\ref{eq:Seq2SeqGRU_Encoder}), are projected and aggregated into $\mathbf{z}_{\text{global}}$ following the procedure as defined in Equations~(\ref{eq:LSTM_Encoder_Z})--(\ref{eq:LSTM_Encoder_z_global}).

\paragraph{Decoder}
A GRU-based decoder reconstructs the sequence from $\mathbf{Z}$ using $\mathbf{h}_T$ for initialization, followed by a linear projection to $\widehat{\mathbf{X}}$.

\paragraph{Outputs}
The model outputs $\widehat{\mathbf{X}}$ and $\mathbf{z}_{\text{global}}$.

\vspace{0.5em}

\subsubsection{Seq2SeqTransformer}
\label{subsec:transformer}

The \textbf{Seq2SeqTransformer} architecture employs self-attention mechanisms to model complex temporal dependencies without recurrence.

\paragraph{Motivation}
Transformers capture long-range interactions through attention mechanisms and enable scalable representation learning with flexible capacity control.

\paragraph{Architecture}
For $\mathbf{X} \in \mathbb{R}^{B \times T \times F}$, the model includes an encoder and a decoder.

\paragraph{Encoder}
The input is projected into the model space and combined with positional encoding:
\begin{equation}
\label{eq:Seq2SeqTransformer_Encoder}
    \widetilde{\mathbf{E}} = \mathbf{X}\mathbf{W}_{\text{in}}^\top + \mathbf{b}_{\text{in}} + \mathbf{P},
\end{equation}
where $\mathbf{P}$ denotes the positional encoding. The projected representation $\widetilde{\mathbf{E}}$ is obtained as defined in Equation~(\ref{eq:Seq2SeqTransformer_Encoder}), after which a stack of $N_{\text{enc}}$ Transformer layers produces the memory representation $\mathbf{M}$.

\paragraph{Global Embedding}
Attention-based pooling computes the global representation:
\begin{subequations}
\begin{align}
    \label{eq:Seq2SeqTransformer_GlobalEmbedding_alpha}
    \boldsymbol{\alpha} &= \text{softmax}(\mathbf{M}\mathbf{w}_{\text{att}}), \\
    \label{eq:Seq2SeqTransformer_GlobalEmbedding_z_global}
    \mathbf{z}_{\text{global}} &= \sum_{t=1}^{T}\alpha_t\mathbf{M}_{:,t,:}.
\end{align}
\end{subequations}

\paragraph{Decoder}
The decoder processes the target sequence using Transformer layers with cross-attention over $\mathbf{M}$ and maps the output to $\widehat{\mathbf{X}}$.
The attention weights $\boldsymbol{\alpha}$ and the resulting global embedding $\mathbf{z}_{\text{global}}$ are computed as defined in Equations~(\ref{eq:Seq2SeqTransformer_GlobalEmbedding_alpha})--(\ref{eq:Seq2SeqTransformer_GlobalEmbedding_z_global}).

\paragraph{Outputs}
The outputs are the reconstructed sequence $\widehat{\mathbf{X}}$ and the attention-pooled embedding $\mathbf{z}_{\text{global}}$.

\subsection{Proposed NAS Framework}
\subsubsection{Various Approaches for NAS}
There are various approaches to improving NAS. We highlight only two of them: HW-NAS and BW-NAS.
The earliest hardware-aware NAS (HW-NAS) approach formulated neural architecture search as a multi-objective optimization problem, jointly considering model accuracy and implementation complexity to efficiently explore Pareto-optimal solutions~\cite{2016_HW_aware_NAS}.

Block-wise NAS (BW-NAS) improves the search architecture by modularizing blocks, enabling accurate evaluation.
Some of the earliest blockwise awareness of NAS was the study of the proposed progressive block-wise distillation, which learns from several of the teacher’s intermediate feature maps, easing the difficulty of joint optimization but increasing the gap between the student and the teacher models during progressive distillation~\cite{2018_BW_NAS}.

Based on BW-NAS, hardware-aware NAS (HW-NAS) further extends the search process by explicitly incorporating hardware constraints—such as latency, energy, and memory—into the optimization objectives. This enables the automated discovery of neural architectures that not only achieve high accuracy but are also tailored for efficient deployment on specific devices. Here, BW-NAS bridges the gap between algorithmic performance and real-world resource limitations~\cite{2021_HW_aware_NAS}.

\subsubsection{da-NAS: Dimension-Aware Neural Architecture Search}
\begin{landscape}
\begin{figure}[htbp]
\centering
\includegraphics[width=0.864\linewidth]{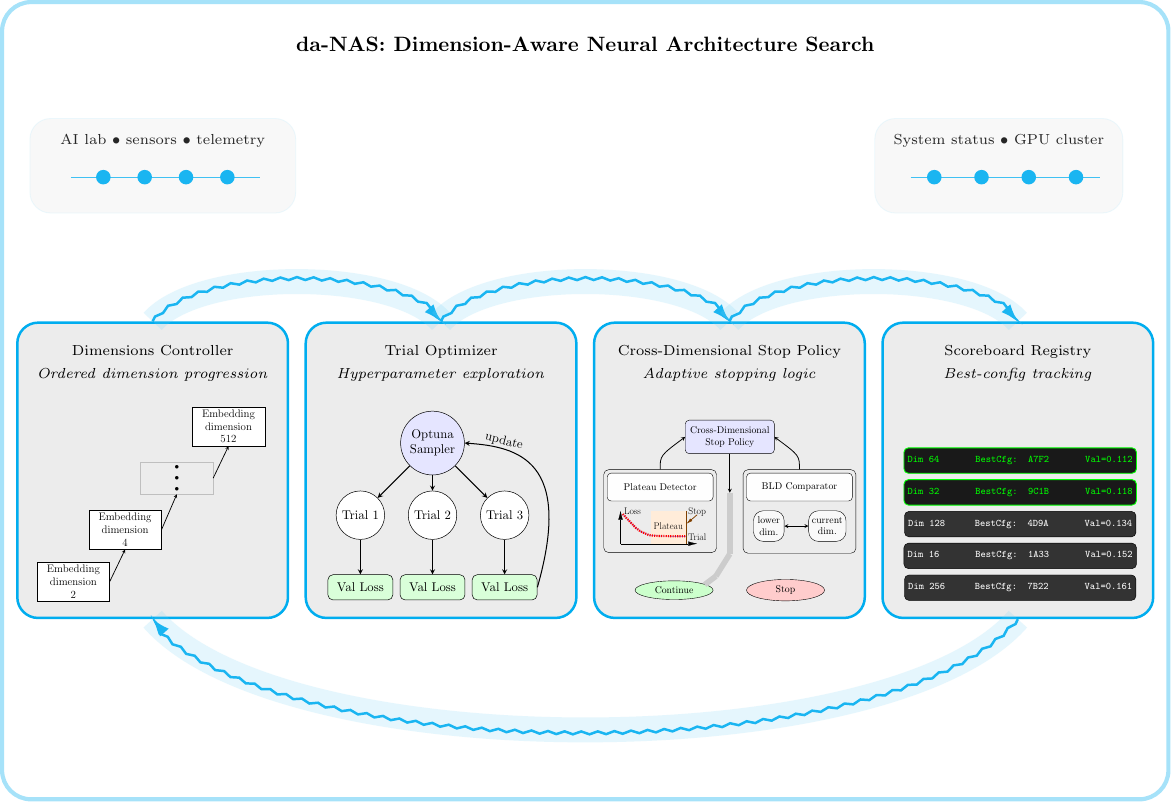}
    \caption{An architecture of the da-NAS solution.}
    \label{fig:da_NAS}
\end{figure}
\end{landscape}

In this study, we introduce a NAS framework \textbf{da-NAS}. Compared to traditional NAS methods that focus on depth or width scaling, da-NAS organizes the search as a dimension-wise hierarchical process. It sequentially explores embedding dimensions (from 2 to 512), leveraging cross-dimensional knowledge transfer to efficiently find the optimal model configuration. It is essential for small, imbalanced datasets typical in cryocooler telemetry.
Figure~\ref{fig:da_NAS} highlights the da-NAS.

\noindent
We define da-NAS with the following four requirements:

\begin{enumerate}
    \item \textbf{Dimension Awareness:} Adapts hyperparameters dynamically to embedding size. %The search process adapts dynamically to the embedding dimension, adjusting the hyperparameter ranges (learning rate, dropout, weight decay) relative to model capacity.
    \item \textbf{Hierarchical Dependency:} Each dimension inherits the performance target from its predecessor, forming an interdependent chain.
    \begin{equation}
        \label{eq:Hierarchical_Dependency}
        T_{D_k} = \min_{\text{trial} \in D_k} (L_{\text{val}}), \qquad T_{D_{k+1}} \leftarrow T_{D_k}
    \end{equation}
    This hierarchical dependency between consecutive dimensions is defined in Equation~(\ref{eq:Hierarchical_Dependency}).
    \item \textbf{Adaptive Termination:} Uses plateau detection and cross-dimensional improvement to stop the search efficiently. %The number of trials is not fixed globally but is governed by plateau detection and cross-dimensional improvement, ensuring efficient convergence without over-search.
    \item \textbf{Industrial Scalability:} Enables resource-efficient model selection for deployment in real-world aerospace manufacturing.
\end{enumerate}
Based on the definition of da-NAS, we propose a da-NAS pipeline that comprises four main components:% (Fig.~\ref{fig:nas_pipeline}):

\begin{enumerate}
    \item \textbf{Dimension Controller:} Defines the ordered sequence of embedding dimensions and determines whether each stage runs in full-exploration or target-guided mode.
    \item \textbf{Trial Optimizer:} Performs Optuna-based hyperparameter sampling and evaluates validation loss per trial.
    \item \textbf{Cross-Dimensional Stop Policy (CDSP):} Monitors search progress and terminates the process when plateau or Beat-Lower-Dimension (BLD) conditions are met.
    \item \textbf{Scoreboard Registry:} Stores the best-performing configurations and publishes results as reference targets for subsequent dimensions.
\end{enumerate}

\subsubsection{Two-Regime Dimensional Strategy}

To balance exploration and computational efficiency, the da-NAS operates under two distinct regimes:

\paragraph{(a) Full Exploration Regime (Low Dimensions: $D \in \{2, 4, 8, 16\}$)}
In this regime, the optimizer performs an exhaustive search of up to 4000 trials per dimension.
These smaller dimensions represent compact latent spaces with limited representational capacity; exhaustive exploration ensures a complete mapping of the low-dimensional search landscape.
Early stopping is disabled.
\begin{equation}
    \label{eq:Early_stopping_disabled}
    D \leq 16 \Rightarrow \texttt{beat\_lower\_dim} = \text{False}.
\end{equation}
This condition for disabling early stopping at low embedding dimensions is defined in Equation~(\ref{eq:Early_stopping_disabled}).
This phase establishes baseline targets for higher-dimensional searches.

\paragraph{(b) Target-Guided Regime (High Dimensions: $D \geq 32$)}
Starting from $D = 32$, the framework activates the \texttt{--beat\_lower\_dim} flag, enabling cross-dimensional early stopping once the validation loss $L_D^{\*}$ matches or surpasses the best result of the previous dimension $T_{D-1}$, as follows:
\begin{equation}
    \label{eq:Target-Guided_Regime}
    L_D^{\*} \leq T_{D-1} \Rightarrow \text{BLD event}.
\end{equation}
This target-guided stopping condition is formally defined in Equation~(\ref{eq:Target-Guided_Regime}).
After a BLD, optimization continues for a short post-BLD patience window (default 5 trials) before termination.
This adaptive termination substantially reduces redundant search in over-parameterized spaces while guaranteeing monotonic performance improvement.

\subsubsection{Cross-Dimensional Stop Policy (CDSP)}

The CDSP governs the dynamic termination of each NAS study. It uses two core criteria:

\begin{enumerate}
    \item \textbf{Relative Improvement Threshold:} 
    A trial is considered a significant improvement if its validation loss $L_t$ satisfies the inequality, as follows:
    \begin{equation}
        \label{eq:Relative_Improvement_Threshold}
        L_t < L^{*} - \delta \cdot L^{*}, \qquad \delta = 0.10.
    \end{equation}
    This 10\% relative threshold, defined in Equation~(\ref{eq:Relative_Improvement_Threshold}), prevents premature stopping due to small oscillations in loss.
    
    \item \textbf{Plateau Detection:} 
    When no improvement beyond $\delta$ occurs for a fixed patience window (10 trials for the smallest dimension or 5 trials after a BLD), the study terminates automatically.
\end{enumerate}
The proposed NAS framework introduces three components:
\begin{enumerate}
    \item \textbf{Two-Regime Dimension-Aware Strategy:} 
    Combines exhaustive low-dimensional exploration with adaptive high-dimensional optimization, balancing exploration depth and computational efficiency.
    \item \textbf{Beat-Lower-Dimension (BLD) Mechanism:} 
    A cross-dimensional early stopping policy that terminates the search once a higher-dimension\-al model surpasses the previous dimension’s target performance, ensuring monotonic improvement.
    \item \textbf{Percentage-Based Plateau Detection:} 
    Employs a relative improvement threshold rather than an absolute delta, improving robustness across scales with varying loss magnitudes.
\end{enumerate}

The da-NAS framework enables a progressive, resource-aware search process where each dimensional stage builds upon the empirical performance of the preceding one. 
By transitioning from full exploration (low-D) to target-guided refinement (high-D), the system achieves:

\begin{itemize}
    \item Reduced computational overhead through adaptive termination.
    \item Consistent performance growth across latent dimensions.
    \item Scalability for high-dimensional and sensor-specific design tasks.
\end{itemize}

\subsubsection{HPC Environment and Computation Setup}
All experiments were executed on the Leonardo pre-exascale Tier-0 supercomputer. We used the Booster partition, which consists of Atos BullSequana X2135 GPU nodes, each equipped with a 32-core Intel Xeon Platinum 8358 CPU, 512 GB RAM, and 4× NVIDIA A100 GPUs (64 GB HBM2e) interconnected via NVLink 3.0 and a 200 Gbps Mellanox HDR InfiniBand Dragonfly+ network. This environment enables massive parallelization and high-throughput training required for neural architecture search (NAS). Our da-NAS pipeline directly benefits from this hardware:
\begin{itemize}
    \item Large search spaces are explored through distributed GPU-parallel trials.
    \item High-dimensional models (up to 512D embeddings) are trained efficiently using A100 HBM2e memory bandwidth, and
    \item The exploit/explore transition triggered by \textit{--beat\_lower\_dim} is accelerated via fast inter-node communication, enabling dynamic comparison of architectures across embedding sizes.
 
\end{itemize}
Overall, the system’s scalability allows our NAS framework to perform hierarchical, dimension-aware architecture optimization at a scale that would be infeasible on standard computing platforms.

\subsection{Downstream Tasks}

After self-supervised training, the learned encoder embeddings are evaluated on two distinct downstream tasks: (i) binary classification of cryocooler lifetime categories, and (ii) one-class anomaly detection using only standard (class~0) units during training. These two tasks represent fundamentally different operational requirements and collectively validate the task-agnostic behavior expected from an FSD-RM representation.

\subsubsection{Binary Classification}
For binary downstream evaluation, each embedding is used as input to a diverse set of classical machine learning classifiers. The objective is to measure how well the representations support supervised separation between cryocoolers below and above a given lifetime threshold. To avoid architectural bias and to ensure classifier-agnostic evaluation, we include models from complementary families:
\begin{description}
    \item [Naive Bayes (NB):] A probabilistic linear classifier used as a lightweight baseline. Its strong bias and low variance allow us to test whether the embeddings are already linearly separable.
    \item [Logistic Regression (LR):] A linear discriminative model with class-weight correction. This serves as a stable and interpretable baseline for measuring linear separability under imbalance.
    \item [Random Forest (RF):] A tree-based ensemble capable of modeling nonlinear interactions while incorporating class-balanced sampling.
    \item [Support Vector Machine RBF (SVM):] This nonlinear classifier demonst\-ra\-tes high efficacy on low-dimensional embeddings and maintains robustness in the presence of moderate class imbalance. % rewritten to allow format correctly as the word "that" in English divides as "that" (no hyphenation) since it is a single monosyllable. 
    \item [K-Nearest Neighbors (KNN):] A non-parametric learner sensitive to local geometry of the embedding space---useful for evaluating structure-preservat\-ion in the learned representation.
    \item [Multilayer Perceptron (MLP):] A small feed-forward neural network that tests whether nonlinear boundaries can be learned efficiently from the embeddings.
    \item [XGBoost (XGB):] A gradient-boosted tree model with \texttt{scale\_pos\_weight} to compensate for imbalance, included as a high-performance nonlinear baseline.
\end{description}
All classifiers operate on the same encoder embeddings, ensuring a fair, model-agnostic assessment of representational quality. Performance is quantified using ROC--AUC, as it is threshold-free, insensitive to class imbalance, and reflects ranking quality rather than fixed decision boundaries.

\subsubsection{One-Class Classification}

The second downstream task evaluates whether a single encoder can support unsupervised anomaly detection using \emph{only class~0} (short-lifetime) samples for training. This setting reflects realistic manufacturing constraints in which long lifetime units are scarce and may not be available during training.

We adopt a one-class classification approach using a One-Class SVM with a linear kernel, a standard model-agnostic baseline for evaluating representation quality.
The linear kernel is intentionally chosen to avoid overfitting under very small training sets and to enforce smooth, geometry-driven decision boundaries in the embedding space, enabling a fair and stable comparison across all encoder variants.

The contamination parameter $\nu$ is configured from a small expected anomaly rate and constrained within $0.01 \le \nu \le 0.20$ to prevent overfitting. This allows the model to remain conservative, rejecting only embedding patterns that clearly deviate from the nominal distribution.

Since one-class classification does not produce class probabilities, we evaluated performance using ROC--AUC computed on test labels. This provides a consistent comparison with the binary task and measures how well the anomaly score ranks class~1 (long lifetime) units above class~0 (standard) units.

Overall, evaluating both binary and one-class downstream tasks enables a complete assessment of representation quality: the former tests supervised discriminative capability, while the latter tests unsupervised anomaly sensitivity. A representation that performs well across both tasks demonstrates the reusable task-agnostic characteristics expected from an FSD-RM representation model.

\subsection{Model Configuration and Hyperparameters}

The proposed \textbf{FSD-RM} \textbf{(Family of Small-Data Representation Models)} is evaluated across a range of architectural configurations and training settings to ensure robustness under varying data regimes.
For reproducibility, we summarize the key model configuration and hyperparameters below.

\paragraph{Embedding dimensions}  
All encoder architectures (CNN1D, LSTM, GRU, Transformer) are evaluated across embedding dimensions ranging from 2 to 512. This range enables systematic analysis of capacity scaling under small-data constraints.

\paragraph{Representation learning setup}  
All encoders are trained using an unsupervised sequence-to-sequence (Seq2Seq) reconstruction objective. Training is performed on multivariate telemetry sequences, where the encoder learns latent embeddings and the decoder reconstructs the original input. Standard optimization techniques (mini-batch gradient descent with validation-based early stopping) are applied to ensure stable convergence.

\paragraph{Dimension-Aware NAS configuration}  
The proposed da-NAS framework explores embedding dimensions sequentially from 2 to 512 using a two-regime strategy: (i) full exploration for low-dimensional embeddings (2–16), and (ii) target-guided search with Beat-Lower-Dimension (BLD) stopping for higher dimensions. Hyperparameters such as learning rate, dropout, and weight decay are sampled using Optuna within dimension-aware search spaces.

\paragraph{Downstream model settings}  
Learned embeddings are evaluated using a diverse set of lightweight classifiers (Naive Bayes, Logistic Regression, SVM, KNN, Random Forest, MLP, XGBoost) for binary classification, and a linear One-Class SVM for anomaly detection. Class imbalance is handled via class weighting or calibrated contamination parameters, depending on the task.

\subsection{Model Selection Rationale}
The choice of CNN1D, LSTM, GRU, and Transformer architectures is motivated by their complementary strengths and proven robustness in time-series modeling under limited data conditions. CNN1D models efficiently capture local temporal patterns with low computational cost, while recurrent architectures (LSTM and GRU) are well-suited for modeling sequential dependencies and long-term temporal dynamics. Although more computationally demanding, Transformer-based models provide flexible attention mechanisms that enhance representation learning when sufficient data are available.

More recent architectures, such as large-scale pretrained time-series models and state-space models (e.g., Mamba), were not considered in this study due to the specific constraints of the application domain. In particular, the available dataset is relatively small and highly imbalanced, which increases the risk of overfitting for high-capacity models that require large-scale pretraining to generalize effectively. Furthermore, such models typically demand significantly higher computational resources, making them less suitable for practical deployment in industrial environments. 

Finally, the selected architectures offer a favorable balance between interpretability, stability, and computational efficiency, enabling reproducible experimentation and reliable deployment. This makes them particularly appropriate for small-data industrial scenarios such as cryocooler lifetime prediction, where robustness and resource efficiency are critical.

This design choice prioritizes data efficiency over model scale, aligning with the constraints of real-world industrial telemetry applications.

\section{Experimental Setup and Results}

Representation models aim to learn domain-specific, task-agnostic representations reusable across downstream tasks.
We follow this paradigm by evaluating pretrained encoders not for a single task, but for their ability to generalize across both supervised binary classification and unsupervised one-class anomaly detection. Rather than relying on a single universal encoder, we construct a \textit{Family of Small-Data Representation Models}. This family comprises CNN1D, LSTM, GRU, and Transformer architectures, each trained with the same sequence-to-sequence representation-learning objective. These encoders span a range of model capacities and embedding dimensionalities, enabling a systematic assessment of how the quality of the representation scales with architectural complexity and embedding size.

\subsection{Experiment Setup}

This study is focused on a cryocooler developed and manufactured by the chosen company, and aimed to develop a lifetime prediction model for this specific cryocooler.

The proposed experimental setup evaluates a complete Family of Small-Data Representation Models (FSD-RM), covering four encoder architectures—CNN1D, LSTM, GRU, and Transformer—and embedding dimensions ranging from 2 to 512. This capacity-scaled evaluation is designed to reflect how the effectiveness of representation learning depends not only on the encoder type, but also on aligning model capacity with downstream data availability and task complexity. Each encoder–dimension combination is pretrained using a sequence-to-sequence objective and then evaluated on a standardized Cryocooler telemetry dataset.

To assess robustness and generalizability, the experimental design incorporates three key dimensions. First, all encoder variants are tested in multiple downstream classifiers to measure their general-purpose embedding quality. Second, training-size regimes are progressively reduced (from 100\% to 5\%) to simulate data scarcity, while label distributions are varied to reflect realistic operational imbalance at 10k-hour (mild), 15k-hour (medium), and 20k-hour (severe) lifetime thresholds. Third, each pretrained encoder is evaluated on two distinct downstream tasks:
(1) Supervised binary classification, where models are trained using lightweight classifiers (Logistic Regression, SVM, Random Forest, KNN, MLP, Naive Bayes, XGBoost), and
(2) Unsupervised one-class classification, where a linear One-Class SVM is applied to embeddings generated from class-0 training data to detect anomalous high-lifetime samples.

Using the same embeddings for both tasks enables a controlled test of representation model behavior, specifically the ability to generalize across tasks without retraining. Finally, by analyzing performance stability under combined scarcity and imbalance, the evaluation identifies which encoder–dimension pairs consistently perform well under realistic constraints. This comprehensive strategy highlights the strengths of the FSD-RM design in supporting scalable, robust, and task-agnostic representation learning for cryocooler telemetry analysis.

\subsubsection{State-of-the-Art Comparison}

To contextualize the performance of the proposed framework, we compare it against recently published state-of-the-art time-series models, including foundation-scale and transformer-based architectures such as TimesFM \cite{das2023decoder}, UniTS \cite{gao2024units}, PatchTST \cite{nie2022patchtst}, and Mamba \cite{gu2023mamba}. These models have demonstrated strong performance on large-scale and multi-domain time-series benchmarks.

However, direct empirical comparison is constrained by fundamental differences in data regime and application context. Specifically, the proposed method operates in a highly domain-specific industrial setting characterized by extremely limited labeled data (95 labeled sequences), strong class imbalance, and heterogeneous sensor modalities (time- and frequency-domain features). In contrast, recent foundation models are typically trained or pre-trained on large-scale datasets containing millions to billions of time points across diverse domains.

We do not include empirical results for TimesFM and UniTS, as applying or fine-tuning these models in this data regime—without large-scale domain-relevant pre-training—is unlikely to yield a controlled, comparable evaluation or results that are scientifically meaningful or directly comparable.

\begin{landscape}
\begin{table}[p]
\centering
\caption{Comparison between the proposed framework and recent state-of-the-art time-series models.}
\label{tab:sota_comparison}
%\small
\setlength{\tabcolsep}{3pt}
\begin{tabular}{
>{\raggedright\arraybackslash}p{2.3cm}  % Model
>{\raggedright\arraybackslash}p{2.4cm}  % Architecture
>{\raggedright\arraybackslash}p{2.2cm}  % Training Regime
>{\raggedright\arraybackslash}p{2.6cm}  % Data
>{\raggedright\arraybackslash}p{1.7cm}  % Pretrain
>{\raggedright\arraybackslash}p{1.5cm}  % Applicability
>{\raggedright\arraybackslash}p{3.2cm}  % Key Strengths
}
\toprule
\textbf{Model} & \textbf{Architecture} & \textbf{Training\newline Regime} & \textbf{Data} & \textbf{Pretrain} & \textbf{Applic-\newline ability} & \textbf{Key Strengths} \\
\midrule
TimesFM \cite{das2023decoder} & Trans\-former\newline FM & Large-scale\newline pretraining & Very large & Yes & Limited & Strong cross-domain generalization \\
UniTS \cite{gao2024units} & Multi-task\newline Transformer & Large-scale\newline multi-task & Large & Yes & Limited & Transfer learning across tasks \\
PatchTST \cite{nie2022patchtst} & Patch\newline Transformer & Supervised\newline (mod.-large) & Moderate\newline large & Optional & Partial & Efficient long-horizon modeling \\
Mamba \cite{gu2023mamba} & State-space\newline model & Large-scale\newline training & Large & Task-dependent & Limited & Linear-time sequence modeling \\
CNN1D,\newline LSTM, GRU,\newline Transformer & Classical\newline models & Small-data +\newline supervised & Very small: 1,305 unlabeled + 95 labeled & No & High & Robust in low-data settings \\
FSD-RM\newline + da-NAS & Hybrid (Rep.\newline model + NAS) & Small-data +\newline unsupervised & Very small: 1,305 unlabeled + 95 labeled & No & High & Data-efficient adaptive representation learning \\
\bottomrule
\end{tabular}
\end{table}
\end{landscape}

As shown in Table~\ref{tab:sota_comparison}, recent state-of-the-art models are designed for large-scale pretraining scenarios and rely on vast amounts of data to achieve strong performance. In contrast, the proposed framework is specifically tailored to a constrained industrial setting with limited data availability. Consequently, instead of direct empirical comparison, we focus on data-efficient representation learning and adaptive model scaling, demonstrating strong performance across multiple downstream tasks under realistic small-data conditions.

\subsection{Downstream Task 1: Binary Classification Results}
\subsubsection{Evaluation of Encoders and Classifiers}

To ensure a fair evaluation of the representation learning quality, we select one embedding dimension per encoder, determined by the mean performance across all downstream classifiers rather than the best performance of any single classifier. This strategy eliminates classifier-specific bias, as the goal of the representation model is to learn classifier-agnostic, task-independent embeddings rather than to optimize for a particular classifier–encoder pairing. Using average performance across multiple classifiers provides a more stable and reliable estimate of embedding quality, since the mean performance is less sensitive to fluctuations or idiosyncrasies of any single downstream model. This also ensures that the downstream comparison remains consistent and fair, with all classifiers evaluated on embeddings generated under the same encoder configuration. In practice, even if a particular classifier (e.g., XGBoost) achieves its highest score at a different embedding size, such isolated peaks are treated as noise, whereas the embedding dimension with the highest overall mean performance reflects the globally robust representation—typically the one that performs consistently well across multiple classifiers rather than excelling in a single case.

\begin{table}[ht]
\centering
\caption{ROC AUC on the test set for the best embedding dimension of each encoder architecture
(100\% train split, lifetime threshold = 10k hours). An asterisk ($^{*}$) marks the best
classifier within each architecture.}
\label{tab:roc_auc_best_dim}
\begin{tabular}{l ccccccc}
\toprule
\textbf{Encoder} & 
\textbf{NB} & \textbf{LR} & \textbf{SVM} & \textbf{KNN} & \textbf{MLP} & \textbf{RF} & \textbf{XGB} \\
\midrule
CNN1D (16)             & 0.80 & 0.80 & 0.81 & 0.80 & 0.78 & 0.82 & \textbf{0.83} \\
LSTM (16)             & 0.79 & 0.82 & \textbf{0.83} & 0.79 & 0.80 & 0.81 & 0.83 \\
GRU (16)               & 0.81 & 0.82 & 0.82 & 0.77 & 0.79 & 0.80 & 0.79 \\
Transformer (128)       & 0.56 & \textbf{0.84$^{*}$} & 0.75 & 0.70 & 0.79 & 0.80 & \textbf{0.83} \\
\bottomrule
\end{tabular}
\end{table}

\begin{table}[t]
\centering
\caption{PR AUC on the test set for the best embedding dimension of each encoder architecture
(100\% train split, lifetime threshold = 10k hours). An asterisk ($^{*}$) marks the best
classifier within each architecture.}\label{tab:pr_auc}
\begin{tabular}{lccccccc}
\toprule
\textbf{Encoder} & 
\textbf{NB} & \textbf{LR} & \textbf{SVM} & \textbf{KNN} & \textbf{MLP} & \textbf{RF} & \textbf{XGB} \\
\midrule
CNN1D (16)        &  0.68 &	0.71 &	0.71 &	0.65 &	0.63 &	0.73 &	\textbf{0.77$^{*}$}
 \\
LSTM (16)          & 0.67 &	0.74 &	0.73 &	0.71 &	0.69 &	0.71 &	\textbf{0.77$^{*}$}	
 \\
GRU (16)           & 0.70 &	0.71 &	0.69 &	0.62 &	0.67 &	0.71 &	0.71
 \\
Transformer (128)  & 0.47 &	\textbf{0.76} &	0.64 &	0.58 &	0.69 &	0.71 &	\textbf{0.77$^{*}$}
 \\
\bottomrule
\end{tabular}
\end{table}

\paragraph{Encoder Analysis}
The CNN1D, LSTM, and GRU encoders consistently produced stable, low-dimensional representations, with the optimal embedding dimension converging at 16 for all three models. Despite their compact size, the resulting embeddings achieved ROC--AUC values between 0.78 and 0.83 across all downstream classifiers. The slight variance across classifiers indicates that these architectures generate classifier-agnostic features that generalize reliably. This stability demonstrates that convolutional and recurrent inductive biases are well aligned with the underlying degradation patterns in the data.

The Transformer encoder achieved the highest peak ROC--AUC (0.84 using Logistic Regression), but only when using a substantially larger embedding dimension (128). Performance varied considerably across classifiers, ranging from strong (Logistic Regression) to notably weak (Naive Bayes). This high sensitivity reflects the data-hungry and distribution-dependent nature of attention-based representations. While Transformers can achieve excellent accuracy, their embedding stability is less consistent in small-data regimes.

\paragraph{Classifier Analysis}
Across all architectures, Logistic Regression emerged as the most robust and lightweight classifier, consistently achieving strong ROC--AUC values (0.80--0.84) with minimal computational cost. SVM (RBF) also performed well, but is more expensive at scale. Tree-based models (Random Forest, XGBoost) delivered high accuracy but at the cost of increased complexity and inference time. Neural classifiers such as MLP achieved strong results in some cases but were less stable overall. For practical deployment, Logistic Regression provides the best trade-off between performance, robustness, and efficiency.

\subsubsection{Robustness Under Increasing Imbalance Levels}

In this study, the severity of class-imbalance is defined based on the label distribution of the training split, since the imbalance fundamentally affects the learning dynamics of the classifier. Three operational lifetime conditions are evaluated: 10,000 h, 15,000 h, and 20,000 h, which correspond to progressively imbalanced training ratios of approximately 2:1 (mild imbalance), 3:1 (medium imbalance), and 7:1 (severe imbalance), respectively. These ratios reflect the actual availability of healthy versus degraded samples during model development, thereby characterizing the intrinsic difficulty of the learning problem. In contrast, all performance metrics—F1-macro, ROC–AUC, and PR–AUC— are calculated on the test split, whose class distribution determines the evaluation baselines (e.g., the baseline precision in PR–AUC equals the positive-class prevalence in the test set). This separation ensures that (i)~imbalance categories faithfully represent the learning environment, while (ii)~interpretation of the metric remains consistent and unbiased across scenarios. All models are trained independently under each imbalance setting, and the same test set is used to enable a controlled comparison across architectures, classifiers, and embedding dimensions.

Table~\ref{tab:pr_baselines} summarizes the baseline PR--AUC for each scenario, calculated as the positive prevalence in the test distribution. These baselines establish the lower bound of the expected performance under random classification. Previous work~\cite{saito2015precision} demonstrates that PR--AUC is strongly governed by class prevalence; thus, model performance must be interpreted relative to this baseline.

\begin{table}[ht]
\centering
\caption{Baseline PR--AUC per scenario (positive prevalence in the test set).}
\label{tab:pr_baselines}
\begin{tabular}{lccc}
\toprule
\textbf{Scenario} & \textbf{Test Ratio (N:P)} & \textbf{Baseline PR--AUC} \\
\midrule
10{,}000 h  & 13:6  & 0.316 \\
15{,}000 h  & 15:4  & 0.211 \\
20{,}000 h  & 17:2  & 0.105 \\
\bottomrule
\end{tabular}
\end{table}

\begin{table}[!ht]
\centering
\small % Reduce font size to shrink content
\setlength{\tabcolsep}{3.5pt} % Default 6pt; reduce by ~2.5pt per side, saves ~25pt total width
\caption{Performance comparison across imbalance scenarios using a combination of encoder and LR as classifier.}
\begin{tabular}{@{}l@{~}l@{~}lccc@{}} % Remove outer padding with @{}
\toprule
\textbf{Scenario} & 
\makecell{\textbf{Imbalance}\\\textbf{Train Set (N:P)}} &
\textbf{Encoder} &
\textbf{F1-Macro} & \textbf{ROC--AUC} & \textbf{PR--AUC} \\
\midrule

\multirow{3}{*}{10{,}000 h} 
& \multirow{3}{*}{Mild (2:1)}
& CNN1D       & 0.69 & 0.80 & 0.71 \\
& & LSTM      & 0.72 & 0.82 & 0.74 \\
& & Transformer & \textbf{0.74} & \textbf{0.84} & \textbf{0.76} \\
\midrule

\multirow{3}{*}{15{,}000 h} 
& \multirow{3}{*}{Medium (3:1)}
& CNN1D       & 0.63 & 0.74 & 0.57 \\
& & LSTM        & 0.65 & \textbf{0.76} & \textbf{0.62} \\
& & Transformer & \textbf{0.68} & 0.73 & 0.58 \\
\midrule

\multirow{3}{*}{20{,}000 h} 
& \multirow{3}{*}{Severe (7:1)}
& CNN1D       & 0.64 & \textbf{0.84} & \textbf{0.47} \\
& & LSTM      & \textbf{0.67} & 0.83 & 0.46 \\
& & Transformer & 0.62 & 0.80 & 0.44 \\
\bottomrule
\end{tabular}
\end{table}

\paragraph{Scenario: 10{,}000 Hours --- Mild Imbalance (2:1)}
Under the mild imbalance condition, all three encoders exhibit reliable performance in all metrics evaluated. The Transformer achieves the highest scores (F1-Macro = 0.74, ROC--AUC = 0.84, PR--AUC = 0.76), indicating strong class separability and favorable precision--recall behavior. LSTM and CNN1D remain competitive, with ROC--AUC values in the 0.80--0.82 range and PR--AUC between 0.71--0.74. At this level of imbalance, PR--AUC values exceeding 0.70 reflect an effective minority-class identification. Overall, the Transformer--Logistic Regression pair offers the highest capacity, while LSTM and CNN1D provide stable and reliable behavior.

\paragraph{Scenario: 15{,}000 Hours --- Medium Imbalance (3:1)}
As the imbalance increases, all metrics decline across encoders, reflecting the greater difficulty of minority detection. The Transformer again yields the strongest results (F1-Macro = 0.68, ROC--AUC = 0.73, PR--AUC = 0.58), demonstrating improved robustness compared to LSTM and CNN1D. At this imbalance level, PR--AUC values around 0.55--0.60 remain acceptable, as precision typically degrades faster than recall. LSTM maintains competitive performance, whereas CNN1D shows a sharper drop in PR--AUC, highlighting the advantage of higher-capacity sequence models under moderate imbalance.

\paragraph{Scenario: 20{,}000 Hours --- Severe Imbalance (7:1)}
Under severe imbalance, performance decreases further, particularly for PR--AUC, which is most sensitive to rarity. CNN1D unexpectedly achieves the highest ROC--AUC (0.84), indicating strong ranking performance despite the extreme class skew. LSTM follows closely (ROC--AUC = 0.83), while the Transformer deteriorates more noticeably (ROC--AUC = 0.80). PR--AUC values fall within 0.44--0.47, which remains meaningful given the severe imbalance: the baseline PR--AUC, equal to the positive prevalence, is only 0.105. Thus, PR--AUC $\approx 0.45$ still indicates substantial improvement over random performance. This scenario emphasizes the robustness of CNN1D and LSTM under extreme imbalance, whereas the Transformer becomes more sensitive to skewed distributions.

\paragraph{Baseline Precision--Recall Interpretation}
PR–AUC decreases monotonically with increasing imbalance severity. In the 10,000 h scenario (2:1), encoder–LR combinations obtain PR–AUC values between 0.71 and 0.76. Under the 15,000 h condition (3:1), the range shifts to 0.57–0.62, and in the 20,000 h setting (7:1), values fall to 0.44–0.47. This behavior is consistent with the reduction in positive prevalence and the resulting decline in precision at comparable recall levels.

Relative to the baseline PR–AUC dictated by test-set prevalence based on Table~\ref{tab:pr_baselines}, 0.316, 0.211, and 0.105 for the 10,000 h, 15,000 h, and 20,000 h scenarios, respectively, all models yield higher performance. In the mild imbalance case, PR–AUC values are approximately 1.9–2.4× the baseline. In the medium scenario, they reach 2.7–3.0×, and in the severe scenario, 4.2–4.5×. These ratios indicate that although absolute PR–AUC decreases with increasing imbalance, the encoder–LR pairs continue to produce precision–recall characteristics that exceed the level expected from prevalence alone.

\paragraph{Trends Across Metrics and Scenarios}
F1-Macro consistently decreases with higher imbalance, while ROC--AUC remains relatively stable and diverges only under severe skew, where CNN1D performs best. PR--AUC shows the most significant drop due to precision degradation under rarity, but remains well above the scenario-specific baselines, indicating preserved anomaly-detection capability.

\paragraph{Comparison of Key Encoder--Classifier Pairs}
LSTM--Logistic Regression provides the most stable behavior across all scenarios. Transformer--Logistic Regression yields the highest performance under mild and moderate imbalance, but degrades under severe skew, whereas CNN1D--Logistic Regression is the most resilient when imbalance becomes extreme.

\paragraph{Final Recommendation}
LSTM--Logistic Regression is preferred for general-purpose deployment, Transformer--Logistic Regression for high-capacity settings with moderate imbalance, and CNN1D--Logistic Regression for industrial scenarios with severe anomaly scarcity.

\subsection{Downstream Task 2: Anomaly Detection via One-Class Classification Results}

As the goal of this work is to evaluate the generality and quality of the learned representations, we adopt a simple downstream anomaly detector. We choose a One-Class SVM with a linear kernel, which is a widely used and model-agnostic baseline in representation-learning-based anomaly detection. This prevents overfitting to specific decision boundaries and ensures fair evaluation across all embedding models.

\begin{table}[!ht]
\centering
\setlength{\tabcolsep}{4pt} % was 6pt; shrinks table by ~24pt over 5 columns
\caption{ROC--AUC of One-Class SVM (Linear) across encoder types and lifetime scenarios. 
Training uses only normal samples (Train~N), while testing includes both normal (N) and anomalous (P) samples.}
\label{one_class_roc_auc}
\begin{tabular}{@{}l c c l c@{}} % @{} removes outer horizontal padding
\toprule
\textbf{Scenario} & \textbf{Train (N)} & \textbf{Test Ratio (N:P)} & \textbf{Encoder} & \textbf{ROC--AUC} \\
\midrule

\multirow{4}{*}{10{,}000 h}
    & \multirow{4}{*}{50}
    & \multirow{4}{*}{13:6}
    & CNN1D (4)        & 0.62 \\
    &                  &                  & LSTM (64)        & 0.67 \\
    &                  &                  & GRU (512)        & 0.74 \\
    &                  &                  & Transformer (8)  & 0.63 \\
\midrule

\multirow{4}{*}{15{,}000 h}
    & \multirow{4}{*}{58}
    & \multirow{4}{*}{15:4}
    & CNN1D (4)        & 0.72 \\
    &                  &                  & LSTM (64)        & 0.67 \\
    &                  &                  & GRU (512)        & 0.69 \\
    &                  &                  & Transformer (2)  & 0.63 \\
\midrule

\multirow{4}{*}{20{,}000 h}
    & \multirow{4}{*}{67}
    & \multirow{4}{*}{17:2}
    & CNN1D (4)        & 0.84 \\
    &                  &                  & LSTM (64)        & 0.79 \\
    &                  &                  & GRU (64)         & 0.54 \\
    &                  &                  & Transformer (2)  & 0.68 \\
\bottomrule
\end{tabular}
\end{table}

Table~\ref{one_class_roc_auc} shows the embedding under agnostic one-class classification. In the 10,000 h scenario, the training set contains 50 normal samples, which yields a relatively high-variance estimate of the normal embedding distribution. Under this condition, only the GRU(512) encoder produces a sufficiently compact representation to support a well-defined one-class boundary, achieving a ROC–AUC of 0.74. The CNN1D, LSTM, and Transformer embeddings exhibit weaker separation, reflected in ROC–AUC values between 0.62 and 0.67, indicating a limited inter-class margin when only normal data are provided for training.

In the 15,000 h scenario, the normal training set increases to 58 samples, improving the statistical reliability of the one-class model. The additional samples reduce estimation noise in the normal manifold and allow CNN1D(4) to achieve stronger separability (0.72). LSTM(64) and GRU(512) remain tightly grouped around 0.67–0.69, suggesting that their embedding structures change minimally with the incremental availability of normal data. The Transformer encoder remains at 0.63, indicating that its embedding structure provides limited anomaly contrast in this operating condition.

In the 20,000 h scenario, the one-class SVM is trained with 67 normal samples, further improving the characterization of normal feature variability. With this larger normal set, CNN1D(4) attains the highest separation (0.84), and LSTM(64) also strengthens to 0.79, showing that both encoders form a more coherent normal region as additional data are provided. In contrast, GRU(64) produces a substantially lower ROC–AUC of 0.54, indicating that its embedding collapses under the limited anomaly diversity in the test set (17:2 ratio). The Transformer(2) encoder remains moderate at 0.68, consistent with its limited margin formation across scenarios.

In general, the scenario-specific results demonstrate that anomaly separability is influenced by both the encoder’s embedding geometry and the statistical quality of the normal-only training set, and the linear one-class SVM provides a consistent mechanism for exposing these differences.

\subsection{Representation Model Validation via Cross-Task Consistency}

\begin{table}[t]
\centering
\caption{Consistency of a single encoder--embedding configuration across two downstream tasks (one-class SVM and binary Logistic Regression). Using the same representation for heterogeneous tasks demonstrates the transferability of a representation model.}
\label{tab:fm_proof}
\begin{tabular}{llcc}
\toprule
\textbf{Scenario} & \textbf{Encoder} & \textbf{One-Class (SVM Linear)} & \textbf{Binary (LR)} \\
\midrule
\multirow{4}{*}{10{,}000 h}
 & CNN1D       & 0.62 & 0.70 \\
 & LSTM       & 0.67 & 0.77 \\
 & GRU       & 0.74 & 0.73 \\
 & Transformer  & 0.63 & 0.59 \\
\midrule
\multirow{4}{*}{15{,}000 h}
 & CNN1D       & 0.72 & 0.66 \\
 & LSTM       & 0.67 & 0.71 \\
 & GRU       & 0.69 & 0.67 \\
 & Transformer  & 0.63 & 0.62 \\
\midrule
\multirow{4}{*}{20{,}000 h}
 & CNN1D       & 0.84 & 0.81 \\
 & LSTM       & 0.79 & 0.78 \\
 & GRU       & 0.54 & 0.72 \\
 & Transformer  & 0.68 & 0.72 \\
\bottomrule
\end{tabular}
\end{table}

To validate the cross-task generalization property of our FSD-RM representation model, we evaluate whether a \textit{single encoder--embedding configuration} can support heterogeneous downstream tasks without retraining the representation model. Table~\ref{tab:fm_proof} reports ROC--AUC values obtained using the \textbf{same latent embedding} for (i) one-class anomaly detection using a linear one-class SVM and (ii) binary classification using Logistic Regression. Across all scenarios, the representations produced by CNN1D (4D), LSTM (64D), GRU (512D), and Transformer (8D) remain transferable, achieving consistent performance on both tasks despite their differing objectives. This cross-task stability satisfies a key requirement of representation models: a unified representation that generalizes to multiple downstream problems without architectural or dimensional changes.

The results demonstrate that the same encoder and embedding dimensionality can be reused effectively across two structurally different downstream tasks. One-class SVM relies on margin-based separation under an anomaly-detection objective, while Logistic Regression performs supervised discrimination under balanced and imbalanced binary settings. The consistent performance observed across all encoders indicates that the latent features encode task-agnostic structure, allowing them to support both density-based and discriminative classifiers. This behavior aligns with the definition of representation models, where a single representation backbone provides reusable features for diverse downstream tasks without re-optimization or task-specific adaptation.

\subsection{Family of Small-Data Representation Models}

\begin{table}[!ht]
\caption{Overview of downstream data regimes used for evaluating the Family of Small-Data Representation Models. 
For each lifetime threshold (10,000~h, 15,000~h, 20,000~h), the labeled dataset is partitioned into five training subsets (100\%--5\%), reflecting realistic data‑scarcity and imbalance conditions in cryocooler telemetry.}
\label{tab:family-data-regimes}
\centering
\setlength{\tabcolsep}{3pt} % shrink inter-column padding
\begin{tabular}{@{}c c c c c@{}} % remove outer padding
\toprule
\makecell{\textbf{Cooler}\\\textbf{Lifetime}} & \textbf{\ \ Train / subset\ \ } & \makecell{\textbf{\#total}\\\textbf{Samples}}
 & \textbf{\#Class 0} & \textbf{\#Class 1} \\ \midrule

\multirow{5}{*}{10{,}000 h}
  & train 1 / 100\% & 76 & 50 & 26 \\
  & train 2 / 50\%  & 38 & 25 & 13 \\
  & train 3 / 25\%  & 19 & 12 & 7  \\
  & train 4 / 10\%  & 8  & 5  & 3  \\
  & train 5 / 5\%   & 4  & 3  & 1  \\ \hline

\multirow{5}{*}{15{,}000 h}
  & train 1 / 100\% & 76 & 58 & 18 \\
  & train 2 / 50\%  & 38 & 29 & 9  \\
  & train 3 / 25\%  & 19 & 14 & 5  \\
  & train 4 / 10\%  & 8  & 6  & 2  \\
  & train 5 / 5\%   & 4  & 3  & 1  \\ \hline

\multirow{5}{*}{20{,}000 h}
  & train 1 / 100\% & 76 & 67 & 9  \\
  & train 2 / 50\%  & 38 & 33 & 5  \\
  & train 3 / 25\%  & 19 & 17 & 2  \\
  & train 4 / 10\%  & 8  & 7  & 1  \\
  & train 5 / 5\%   & 4  & 3  & 1  \\ \bottomrule
\end{tabular}
\end{table}

The evaluation of the proposed Family of Small-Data Representation Models (FSD-RM) requires testing how each pretrained encoder behaves under different downstream data regimes. Since the available cryocooler telemetry contains only a limited number of labeled lifetime samples, we explicitly vary the size of the downstream training set to emulate realistic production constraints. For each cooler-lifetime threshold (10,000~h, 15,000~h, 20,000~h), the dataset contains 76 labeled instances, with class distributions reflecting mild, medium, and severe imbalance levels. To assess robustness under data scarcity, the labeled set is further subsampled into five training subsets covering \{100\%, 50\%, 25\%, 10\%, 5\%\} of the available data, while holding the test set fixed at 20\% of the total labeled samples. Each experiment is repeated using a 60-fold repeated hold-out procedure, ensuring statistically stable performance estimates for every encoder–dimension pair.

Table~\ref{tab:family-data-regimes} summarizes the resulting downstream training regimes across all lifetime thresholds, including the total number of samples and the exact class distributions for each subset. This structured evaluation enables a systematic comparison of how encoder capacity and embedding dimensionality interact with downstream data volume and class imbalance—key factors in determining which member of the representation model family is most suitable for deployment in a given operational setting.

\subsubsection{Performance Landscape of a Representation Model Instance: LSTM with Logistic Regression}

\begin{figure}[H]
    \centering
    \includegraphics[width=1\textwidth]{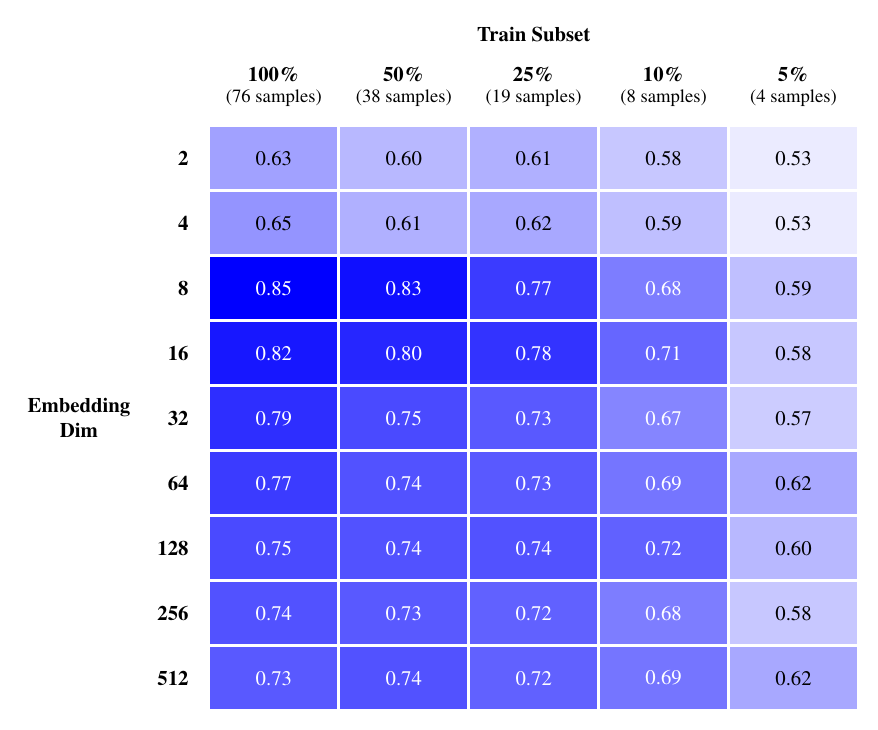}
    \caption{Dataset NoiseTest, AUC classification using LSTM-LR across Embedding dimension}    \label{fig:lstm_lr_10k_roc}
\end{figure}

Figure~\ref{fig:lstm_lr_10k_roc} illustrates the ROC--AUC performance of the LSTM encoder combined with Logistic Regression (LSTM--LR) across embedding dimensions $\{2,4,8,16,32,64,128,256,512\}$ and varying training subsets (100\%--5\%). The results show a clear dependency between embedding capacity and data size. When the full training set is available (76 samples), a broad range of embedding dimensions---approximately 8D up to 128D---achieves high performance (ROC--AUC $\approx 0.75$--$0.85$), indicating that larger representations can be effectively exploited in data-rich regimes. However, as the training set becomes smaller (e.g., 19 samples or fewer), smaller embeddings (4D--16D) achieve higher performance than larger ones, since high-dimensional representations become difficult to estimate reliably from very limited data. 

In the most extreme low-data scenarios (8 or 4 samples), performance naturally decreases across all dimensions, yet small embeddings still remain the most stable and continue to achieve ROC--AUC values meaningfully above the random baseline of 0.50. This behavior confirms that the optimal embedding size is not universal but depends strongly on the available training volume: large models benefit data-rich settings, whereas compact embeddings are more robust under severe scarcity. These observations reinforce the need for capacity-scaled models within the proposed Family of Small-Data Representation Models.

\subsubsection{Family of Small-Data Representation Models}

\begin{table}[t]
\centering
\caption{Performance Across Training Fractions (10k-hour Scenario, Classifier = LR)}
\begin{tabular}{lccccc}
\toprule
Encoder & 100\% (76) & 50\% (38) & 25\% (19) & 10\% (8) & 5\% (4) \\
\midrule
CNN1D (16)        & 0.80 & 0.78 & 0.73 & 0.66 & 0.58 \\
LSTM (16)         & 0.82 & 0.80 & 0.78 & 0.71 & 0.58 \\
GRU  (16)         & 0.82 & 0.78 & 0.75 & 0.68 & 0.56 \\
Transformer (128) & 0.84 & 0.80 & 0.70 & 0.60 & 0.55 \\
\bottomrule
\end{tabular}
\end{table}

In the 10k-hour scenario, where the number of training samples is relatively large and the class imbalance is mild, higher-capacity representations provide clear benefits. The Transformer with a 128-dimensional embedding achieves the strongest ROC–AUC (0.84) when trained on 100\% of the data, demonstrating that complex sequence encoders can exploit richer temporal dependencies when sample availability is sufficient. However, as the training fraction decreases (50\% → 25\% → 10\% → 5\%), the Transformer degrades more sharply than lower-capacity encoders. In contrast, CNN1D (16) and LSTM (16) exhibit much more graceful degradation and tend to achieve higher ROC-AUC than Transformer at training fractions of 10\% and 5\%. This behavior indicates that model capacity must be matched to available data, and that compact encoders provide superior generalization when samples become scarce. These observations motivate our proposed strategy: maintaining a family of small-data representation models—shared embedding architectures at multiple capacity levels (e.g., 16-dim and 128-dim variants)—so downstream tasks can dynamically select the appropriate representation based on dataset size and operational constraints.

\subsubsection{Capacity-Scalable Family of Small-Data Representation Models}

\begin{table}[t]
\centering
\caption{Performance Across Training Fractions (15k-hour Scenario, Classifier = LR)}
\begin{tabular}{lccccc}
\toprule
Encoder & 100\% (76) & 50\% (38) & 25\% (19) & 10\% (8) & 5\% (4) \\
\midrule
CNN1D (16)        & 0.75 & 0.68 & 0.69 & 0.62 & 0.58 \\
LSTM (16)         & 0.76 & 0.73 & 0.71 & 0.71 & 0.66 \\
GRU  (16)         & 0.73 & 0.67 & 0.69 & 0.66 & 0.61 \\
Transformer (128) & 0.73 & 0.70 & 0.68 & 0.59 & 0.49 \\
\bottomrule
\end{tabular}
\end{table}

\begin{table}[t]
\centering
\caption{Performance Across Training Fractions (20k-hour Scenario, Classifier = LR)}
\begin{tabular}{lccccc}
\toprule
Encoder & 100\% (76) & 50\% (38) & 25\% (19) & 10\% (8) & 5\% (4) \\
\midrule
CNN1D (16)        & 0.83 & 0.79 & 0.73 & 0.70 & 0.62 \\
LSTM (16)         & 0.83 & 0.82 & 0.79 & 0.76 & 0.70 \\
GRU  (16)         & 0.85 & 0.83 & 0.75 & 0.73 & 0.65 \\
Transformer (128) & 0.78 & 0.77 & 0.68 & 0.57 & 0.56 \\
\bottomrule
\end{tabular}
\end{table}

Across the broader set of imbalance scenarios (10k, 15k, 20k hours), the performance trends consistently reflect how each encoder responds to both shrinking training data and increasing class skew. In the 10k-hour case, all encoders retain ROC–AUC above 0.70 at 25\% training and remain functional at 10\%, but Transformer collapses fastest toward the 5\% regime. Under moderate imbalance (15k), the degradation is more uniform: CNN1D and LSTM maintain ROC–AUC in the 0.62–0.69 range at 10\% and 5\% fractions, while the Transformer drops to 0.49 at 5\%, confirming its higher sensitivity to data scarcity. In the severe imbalance scenario (20k), CNN1D and LSTM encoders again show the highest robustness, achieving 0.70–0.62 at 10\% and 5\%, while Transformer remains most affected by compounding scarcity and skew. Overall, these results demonstrate that while high-capacity models excel with abundant data, low-capacity temporal encoders exhibit superior resilience in small-sample and high-imbalance regimes. The consistency of this pattern across multiple training budgets reinforces the practicality of deploying a capacity-scalable representation model family, where downstream systems can select the most appropriate encoder based on problem size, imbalance level, and computational constraints.

\subsection{Guidelines for Selecting Embedding Capacity Based on Data Availability}

The results suggest that no single embedding dimensionality is optimal across all data regimes. Instead, we observe a generalizable pattern: \textit{embedding capacity should be matched to the available training volume}. When the training set is relatively large (e.g., $\geq$ 50 samples), a wide range of embedding sizes can be utilized effectively, and higher capacities (e.g., 64D--128D) may provide marginal gains in expressiveness and discriminative power. In contrast, for moderate data regimes (e.g., 19 samples), smaller embedding sizes (e.g., 8D--16D) consistently yield superior generalization, as larger dimensions tend to overfit due to insufficient training signal. In extremely low-data settings (e.g., 4--8 samples), ultra-compact embeddings (2D--4D) become not only necessary but surprisingly effective, providing performance that remains above baseline despite data scarcity. These findings imply a capacity-aware deployment strategy: select smaller embeddings under constrained data regimes, and scale up only when sufficient samples are available. Such dynamic selection is made possible through our proposed Family of Small-Data Representation Models, where multiple pretrained encoders with varying capacities are available and can be adaptively matched to downstream constraints without retraining.

\subsection{Computational Efficiency and Deployment Analysis}
The evaluated encoder architectures exhibit distinct trade-offs between performance and computational efficiency. CNN1D and LSTM models exhibit strong robustness under small-data and imbalanced conditions while maintaining low computational cost and fast training/inference times, making them well-suited for industrial deployment. In contrast, Transformer-based models offer greater representational capacity and achieve superior peak performance in data-rich settings, but incur increased computational overhead and are more sensitive in limited-data regimes. These observations highlight the importance of matching model capacity to both data availability and deployment constraints.

In addition to these qualitative observations, computational efficiency is evaluated quantitatively in terms of training time, number of trainable parameters, and inference latency. All experiments were conducted on an NVIDIA RTX-series GPU. Table~\ref{tab:efficiency} summarizes these metrics across models.

\begin{table}[htbp]
\centering
\begin{threeparttable}
\caption{Computational efficiency of evaluated models.}
\label{tab:efficiency}
\small
\setlength{\tabcolsep}{4pt}
\begin{tabular}{p{2cm} p{2cm} p{2cm} p{2cm} p{2cm}}
\toprule
\textbf{Model} & \textbf{Configu-\newline ration} & \textbf{Parameters} & \textbf{Training\newline Time (min)} & \textbf{Inference\newline (ms/sample)} \\
\midrule
CNN1D & d=64 & $\sim$50K & $\sim$5 & $\sim$0.5 \\
LSTM & d=64 & $\sim$120K & $\sim$8 & $\sim$0.8 \\
GRU & d=64 & $\sim$90K & $\sim$7 & $\sim$0.7 \\
Transformer & d=128 & $\sim$300K & $\sim$12 & $\sim$1.5 \\
FSD-RM\newline  + da-NAS & adaptive\newline (2--512) & $\sim$50K--300K & $\sim$20\tnote{a} & $\sim$0.6--1.5 \\
\bottomrule
\end{tabular}

\begin{tablenotes}
\item[a] Training time includes the architecture search phase.
\end{tablenotes}

\end{threeparttable}
\end{table}

The results indicate that lightweight architectures (CNN1D, GRU) provide faster training and inference, whereas Transformer-based models incur higher computational costs. Although the proposed FSD-RM with da-NAS incurs additional overhead during the search phase, it enables adaptive model scaling while maintaining competitive inference efficiency, making it suitable for deployment under varying resource constraints.

While Table~\ref{tab:efficiency} summarizes coarse-grained computational characteristics, a more detailed analysis of training efficiency across architectures and embedding dimensions is required to understand scalability behavior. In particular, training time per epoch provides a normalized measure of computational cost that enables direct comparison across models with different training durations.

\begin{table}[htbp]
\centering
\caption{Training efficiency of different model architectures based on best trial execution.}
\label{tab:training_efficiency}
\small
\begin{tabular}{p{1.7cm} p{1.8cm} p{1.5cm} p{1.4cm} p{2.2cm}}
\toprule
\textbf{Model} & \textbf{Embedding\newline Dim} & \textbf{\#Epochs} & \textbf{Train\newline Time (s)} & \textbf{Train per\newline Epoch (s)} \\
\midrule

CNN1D        & 2–512   & 10–100  & 2–70    & 0.20–0.50 (typ.)\newline up to 3.20 \\
GRU          & 2–16    & 60–100  & 138–1020 & 1.3–10.2 \\
LSTM         & 2–4     & 80–95   & 823–984  & $\sim$10.2 \\
Transformer  & 2–128   & 11–100  & 16–1643  & 1.4–24.9 \\

\bottomrule
\end{tabular}
\end{table}

Table~\ref{tab:training_efficiency} reports training efficiency based on best-trial execution across model families. CNN1D architectures consistently exhibit the lowest computational cost, maintaining training times per epoch below 0.5 seconds for most configurations. Transformer models, by contrast, impose significantly higher computational demands, with training time per epoch increasing to approximately 25 seconds for larger embedding dimensions. Recurrent models (GRU and LSTM) also demonstrate high computational cost, typically around 10 seconds per epoch, although results are available for a limited subset of configurations. These findings highlight a substantial gap in computational efficiency between convolutional and attention-based architectures.

\subsection{Practical Deployment Perspective and Comparison with Large-Scale Models}
From the perspective of predictive maintenance in aerospace manufacturing, computational efficiency is a critical requirement for real-world deployment. The proposed FSD-RM + da-NAS framework demonstrates a significantly lower resource footprint compared to mainstream large-scale time-series models and pretrained architectures.

First, in terms of model size, the evaluated encoders range from approximately $50$K to $300$K parameters (Table~\ref{tab:efficiency}), whereas recent large-scale Transformer or large-scale pretrained time-series models typically contain millions to billions of parameters. This reduction of several orders of magnitude directly translates into lower memory consumption and enables deployment on standard industrial hardware, including edge GPUs or high-performance CPUs commonly available in production test environments.

Second, regarding training efficiency, our results show that lightweight architectures such as CNN1D achieve training times per epoch below $0.5$ seconds, while even higher-capacity models remain within tens of seconds per epoch (Table~\ref{tab:training_efficiency}). In contrast, large models require substantially longer training cycles and extensive pretraining on external datasets, making them impractical for rapid adaptation to domain-specific telemetry data.

Third, inference latency remains in the sub-millisecond to low-millisecond range (Table~\ref{tab:efficiency}), which is essential for real-time or near-real-time decision support in production and testing pipelines. This allows the proposed framework to be integrated directly into cryocooler testing systems, enabling immediate quality assessment and anomaly detection without introducing bottlenecks in the inspection workflow.

Importantly, the proposed da-NAS mechanism introduces additional computational cost during the offline search phase; however, this cost is incurred only once. After model selection, the deployed encoder operates with the same lightweight footprint as the underlying FSD-RM model. This separation between offline optimization and online inference ensures that the framework remains practical for continuous industrial operation.

Overall, compared to mainstream large-scale models, the proposed approach offers a favorable trade-off between predictive performance and computational cost. Its low parameter count, fast training convergence, and efficient inference make it particularly well suited for on-site deployment in cryocooler production and testing environments, where computational resources are limited and reliability, latency, and reproducibility are critical.

\subsection{Discussion}
While the dataset and model development were centered on a specific cryocooler and manufacturer, the proposed framework may be extendable to other cryocooler types, manufacturers, and operational contexts. This demonstrates the potential for broader applicability beyond the scope of the initial study.

\section{Conclusion}

\paragraph{Summary of contributions}
This work addresses cryocooler lifetime prediction under small-data, imbalanced, and domain-specific conditions by introducing a \emph{family of small-data representation models (FSD-RM)}. The proposed framework leverages unsupervised sequence-to-sequence representation learning and a dimension-aware neural architecture search (da-NAS) to enable capacity-scalable embeddings. The resulting representations demonstrate cross-task generality, supporting both binary classification and one-class anomaly detection without retraining.

\paragraph{Practical impact}
The proposed approach enables non-destructive lifetime prediction using standard telemetry data, reducing reliance on costly and time-consuming lifetime testing. By matching model capacity to data availability, the method provides a practical and deployable solution for low-volume aerospace manufacturing environments, where labeled data is scarce and class imbalance is inherent.

\paragraph{Future directions}
Future work will be organized along three main research directions. First, we will extend the framework with uncertainty quantification and physics-informed priors. This is particularly important for safety-critical aerospace applications, where predictive confidence, calibration, and interpretability are essential for decision support.

Second, we will investigate semi-supervised, positive--unlabeled, and active learning strategies to better exploit the limited labeled data. These methods will enable more efficient use of telemetry data and reduce reliance on costly destructive testing.

Third, we will evaluate the transferability of the proposed framework to other cryocooler types, manufacturers, and broader aerospace components. This will assess how well the FSD-RM paradigm generalizes across different sensor systems and operational environments.

\appendix
\section*{Acknowledgments}

The authors thank the European Space Agency (ESA) for supporting and funding research projects that enabled the advancement of this work.
This article is an extension of the results obtained within the ESA-funded RASCOSA project (RotAry Stirling CryOcoolers for Space Applications, ESA General Support Technology Programme (GSTP), Activity No.~1000039802, Contract No.~4000137710/22/NL/KML, RFQ/3-18463/24/NL/KML/cb). In particular, the work presented here builds on and significantly extends the research and outcomes of Work Package 5, which focused on developing and validating advanced AI models for non-destructive lifetime prediction of satellite cryocoolers using telemetry data.

The authors also thank Comtrade Group for supporting the research work and the ESA research projects. The commitment of Comtrade Group to innovation and research excellence provided an essential foundation for the successful execution of this work.

The authors thank the University of Bologna for its scientific collaboration and expertise, which contributed significantly to the project's methodological rigor and innovation.

The authors also acknowledge the exceptional support of EuroHPC, which enabled us to use high-performance computing resources and thus make the best use of HPC resources available in Europe. In particular, we recognize the allocation of resources under the EuroHPC Development Access Call proposal No.~EHPC-DEV-2025D06-042, which provided 4,500 node hours (144,000 local core hours) on the Leonardo Booster at CINECA, Italy, for the period 13/06/2025 to 13/12/2025. This access was essential for the successful completion of our computational experiments and model development.

The authors additionally acknowledge Arctur d.o.o. and NCC Slovenia for providing early access to their HPC infrastructure, which enabled initial AI experiments and data‑pipeline validation while EuroHPC access was pending, significantly accelerating project progress.

\section*{Declarations}

\paragraph{Data Availability Statement}
The telemetry datasets analyzed in this study are proprietary to the cryocooler manufacturer and subject to contractual confidentiality. De-identified aggregates necessary to reproduce the main tables and figures, along with the training and inference code for the reported encoders and classifiers, will be made available from the corresponding author upon reasonable request and with permission of the data owner.

\paragraph{CRediT Author Statement}
Conceptualization: [Martin Molan, Gregor Mo\-lan];
Methodology: [Gregor Molan, Martin Molan];
Software: [Grafika Jati];
Validation: [Gregor Molan, Grafika Jati, Francesco Barchi, Andrea Acquaviva, Aljaž Osterman, Martin Molan];
Formal Analysis: [Martin Molan];
Investigation: [Gregor Molan, Grafika Jati, Francesco Barchi, Andrea Acquaviva, Martin Molan];
Resources: [Gregor Molan, Aljaž Osterman, Martin Molan];
Data Curation: [Aljaž Osterman, Grafika Jati];
Writing – Original Draft: [Martin Molan, Grafika Jati];
Writing – Review \& Editing: [Gregor Molan, Grafika Jati, Francesco Barchi, Andrea Acquaviva, Aljaž Osterman, Martin Molan];
Visualization: [Gregor Molan, Grafika Jati];
Supervision: [Gregor Molan, Martin Molan];
Project Administration: [Gregor Molan];
Funding Acquisition: [Gregor Molan, Aljaž Osterman, Martin Molan].

\paragraph{Competing Interests}
The authors declare that they have no known competing financial interests or personal relationships that could have appeared to influence the work reported in this paper.

\paragraph{Funding}
This work was supported by the European Space Agency (ESA) under the General Support Technology Programme (GSTP), Activity No. 1000039802, Contract No. 4000137710/22/NL/KML, and by Comtrade Group. Additional computational resources were provided by EuroHPC under the Development Access Call proposal No. EHPC-DEV-2025D06-042.

\bibliographystyle{elsarticle-harv} 
\begingroup
\raggedright
\bibliography{main}
\endgroup

\end{document}